%% file: arxiv.tex
\documentclass[sigconf,screen,authorversion,nonacm]{acmart}

\input{math_commands.tex}

\usepackage{enumitem}
\usepackage{hyperref}
\usepackage{url}
\usepackage{bbm}

\usepackage{graphicx}
\usepackage{multirow}
\usepackage{booktabs}
\usepackage{amsthm}
\theoremstyle{definition}
\newtheorem{definition}{Definition}[section]
\newtheorem{theorem}{Theorem}[section]
\newtheorem{lemma}[theorem]{Lemma}
\usepackage{tikz}
\usepackage{subcaption}

\usepackage{xspace}
\newcommand{\proj}{Bloom Signature\xspace}

\AtBeginDocument{%
  \providecommand\BibTeX{{%
    \normalfont B\kern-0.5em{\scshape i\kern-0.25em b}\kern-0.8em\TeX}}}

\usepackage{fancyhdr}
\AtBeginDocument{%
    \addtolength{\footskip}{2.0\baselineskip}%
    \fancyfoot[L]{\textit{\textbf{Preprint.}}}%
}

\setcopyright{acmcopyright}
\copyrightyear{2018}
\acmYear{2018}
\acmDOI{XXXXXXX.XXXXXXX}

\acmConference[Conference acronym 'XX]{Make sure to enter the correct
  conference title from your rights confirmation email}{June 03--05,
  2018}{Woodstock, NY}
\acmPrice{15.00}
\acmISBN{978-1-4503-XXXX-X/18/06}

\begin{document}

\title{Learning Scalable Structural Representations for Link Prediction with Bloom Signatures}

\author{Tianyi Zhang}
\email{tz21@rice.edu}
\authornote{Both authors contributed equally to this research.}
\affiliation{
  \institution{Rice University}
  \city{Houston}
  \state{TX}
  \country{USA}}

\author{Haoteng Yin}
\email{yinht@purdue.edu}
\authornotemark[1]
\affiliation{
  \institution{Purdue University}
  \city{West Lafayette}
  \state{IN}
  \country{USA}}

\author{Rongzhe Wei}
\email{rongzhe.wei@gatech.edu}
\affiliation{
  \institution{Georgia Tech.}
  \city{Atlanta}
  \state{GA}
  \country{USA}}

\author{Pan Li}
\email{panli@gatech.edu}
\affiliation{
  \institution{Georgia Tech.}
  \city{Atlanta}
  \state{GA}
  \country{USA}}

\author{Anshumali Shrivastava}
\email{anshumali@rice.edu}
\affiliation{
  \institution{Rice University}
  \city{Houston}
  \state{TX}
  \country{USA}
}

\renewcommand{\shortauthors}{Zhang and Yin, et al.}

\begin{abstract}
  \input{abstract}
\end{abstract}



\settopmatter{printfolios=true}

\maketitle

\input{1-Introduction}

\input{2-Preliminary}
\input{3-Methodology}
\input{4-Experiment}
\input{5-Conclusion}

\begin{acks}
The authors would like to thank Beatrice Bevilacqua and Yucheng Zhang for their helpful discussions. Haoteng Yin and Pan Li are supported by the JPMorgan Faculty Award, NSF awards OAC-2117997, IIS-2239565.
\end{acks}

\bibliographystyle{ACM-Reference-Format}
\bibliography{ref}

\appendix
\input{6-Appendix}

\end{document}

%% file: math_commands.tex
\usepackage{amsmath,amsfonts,bm}

\def\eqref#1{equation~\ref{#1}}

\def\1{\bm{1}}

\def\ve{{\bm{e}}}

\def\vh{{\bm{h}}}

\def\vm{{\bm{m}}}

\def\vs{{\bm{s}}}

\def\vu{{\bm{u}}}
\def\vv{{\bm{v}}}

\def\vx{{\bm{x}}}

\DeclareMathAlphabet{\mathsfit}{\encodingdefault}{\sfdefault}{m}{sl}
\SetMathAlphabet{\mathsfit}{bold}{\encodingdefault}{\sfdefault}{bx}{n}



%% file: abstract.tex
Graph neural networks (GNNs) have shown great potential in learning on graphs, but they are known to perform sub-optimally on link prediction tasks. Existing GNNs are primarily designed to learn node-wise representations and usually fail to capture pairwise relations between target nodes, which proves to be crucial for link prediction. Recent works resort to learning more expressive edge-wise representations by enhancing vanilla GNNs with structural features such as labeling tricks and link prediction heuristics, but they suffer from high computational overhead and limited scalability. To tackle this issue, we propose to learn structural link representations by augmenting the message-passing framework of GNNs with Bloom signatures. Bloom signatures are hashing-based compact encodings of node neighborhoods, which can be efficiently merged to recover various types of edge-wise structural features. We further show that any type of neighborhood overlap-based heuristic can be estimated by a neural network that takes Bloom signatures as input. GNNs with Bloom signatures are provably more expressive than vanilla GNNs and also more scalable than existing edge-wise models. Experimental results on five standard link prediction benchmarks show that our proposed model achieves comparable or better performance than existing edge-wise GNN models while being 3-200 $\times$ faster and more memory-efficient for online inference.

%% file: 1-Introduction.tex
\section{Introduction}
Link prediction is one of the fundamental tasks in graph machine learning and has wide real-world applications in network analysis, recommender systems~\cite{bennett2007netflix,zhang2020inductive}, knowledge graph completion~\cite{teru2020inductive,zhu2021neural}, scientific studies on graph-structured data (e.g. protein, drug, blood vessels) in biochemistry~\cite{szklarczyk2019string,stanfield2017drug} and neuroscience~\cite{paetzold2021whole}. Early research \citep{liben2003link} investigated using hand-crafted score functions to measure the neighborhood similarity between two target nodes for estimating the likelihood of an existing link. Representative score functions including Common Neighbor~\cite{barabasi1999emergence}, Jaccard index~\cite{jaccard1901distribution}, Adamic-Adar~\cite{adamic2003friends} and Resource Allocation~\cite{zhou2009predicting} are deterministic and localized, often referred as link prediction heuristics. To incorporate the global structure, node embedding~\cite{perozzi2014deepwalk,grover2016node2vec} or matrix factorization-based~\cite{qiu2018network} approaches are proposed, which map each node in the graph into a low-dimensional latent space. These vectorized node embeddings preserve certain structural properties of the input graph that can be used for link prediction. Recently, graph neural networks (GNNs)~\cite{kipf2016semi,hamilton2017inductive} have begun to dominate representation learning on graphs, due to its benefits in combining node features and graph structures through a message-passing framework~\cite{gilmer2017neural}. To predict a link, GNNs are first applied to the entire graph to obtain the node-wise representation, and then embeddings of two target nodes are aggregated and fed into a classifier to estimate the likelihood. GNNs have shown excellent performance on node-level tasks, but sometimes only achieve subpar performance on link prediction, and may even be worse than unsupervised embedding methods or simple heuristics~\cite{zhang2018link,zhang2021labeling}.

\begin{figure}
    \centering
    \includegraphics[width=0.99\linewidth]{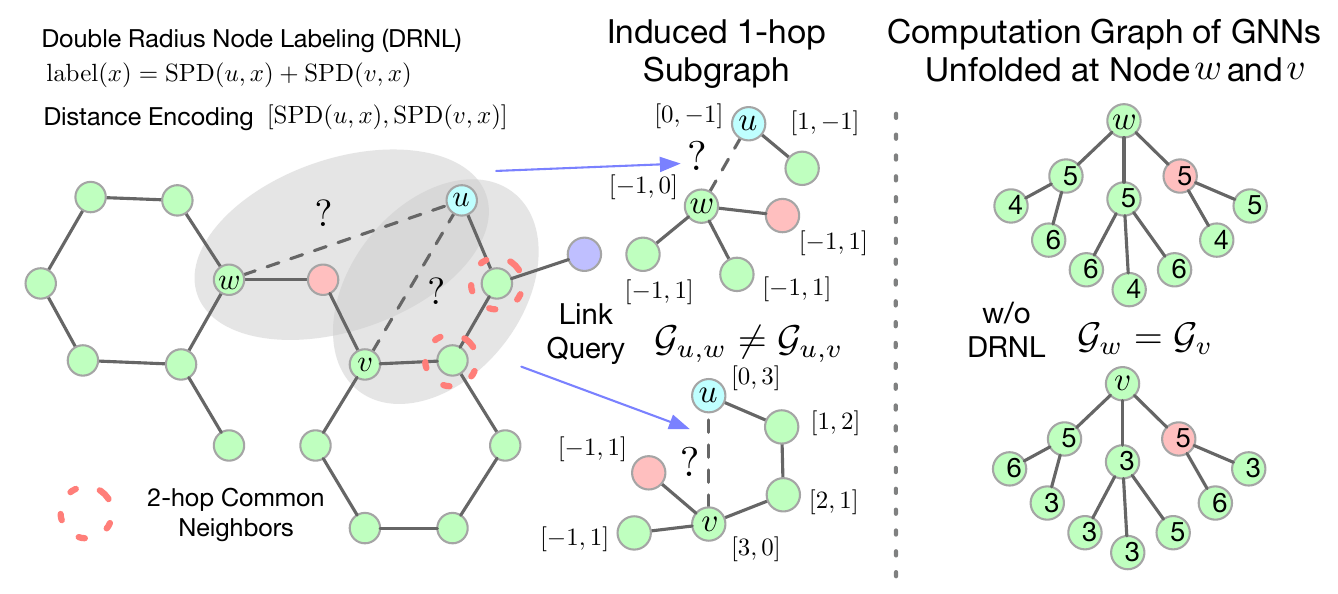}
    \vspace{-5mm}
    \caption{\small{Vanilla GNNs cannot correctly predict whether $u$ is more likely linked with $v$ or $w$: without distinct features, $v$ and $w$ have the same node representation. Representations of subgraphs with proximity-based labels (Left: encoding of shortest path distances up to 3 hops; Right: double radius node labeling) are more expressive to distinguish node pairs $(u,v)$ and $(u,w)$.}}
    \label{fig:expressive}
    \vspace{-7mm}
\end{figure}

Several works have revealed the deficiency of directly aggregating node embeddings produced by GNNs as the representation of a link for prediction~\cite{zhang2018link,srinivasan2019equivalence,li2020distance,zhang2021labeling,chamberlain2022graph,wang2023neural}: (1) It cannot measure the similarity between node neighborhoods as link prediction heuristics do, where~\citet{chen2020can} proves that GNNs are incapable of counting connected substructures like triangles. (2) It cannot distinguish target nodes that are isomorphic within its $k$-hop induced subgraphs (such as nodes $v,w$ in Fig. \ref{fig:expressive}), where~\citet{xu2019powerful} proves that the expressive power of GNNs is bounded by the 1-Weisfeiler-Lehman (1-WL) test. For nodes with the isomorphic neighborhood and no distinct features, due to limited expressiveness, vanilla GNNs map them into the same representation (nodes $v,w$ have the identical unfolded computation tree as depicted in Fig. \ref{fig:expressive} Right). As a node-wise model, GNNs are incapable of producing an effective structural representation of target links, which requires encoding in terms of pairwise relationships of two given nodes.

Recent studies resort to learning more expressive link representations through edge-wise models. Two types of approaches are used to alleviate the limitations of vanilla GNNs in generating link representations by introducing pairwise features. \textbf{Coupled}: labeling tricks are a family of target-link-specific structural features~\cite{zhang2018link,you2021identity,li2020distance,zhu2021neural,zhang2021labeling}. It injects the dependency between nodes in a target link by assigning nodes with separate proximity-based labels in their $k$-hop enclosing subgraphs. These added labels could break the structural symmetry of nodes with isomorphic neighborhoods (see node labels in Fig. \ref{fig:expressive} Right), and thus significantly boost the performance of GNNs on multiple link prediction benchmarks~\cite{hu2020open}. However, subgraph labeling is conditioned on target links, which introduces substantial computation overhead. \textbf{Decoupled}: instead of modifying the input graph with labeling, hand-crafted pairwise features are directly used to generate link representations, such as intersection and difference of high-order neighborhoods~\cite{yun2021neo,chamberlain2022graph} and the geodesic path~\cite{kong2022geodesic} between nodes in a target link. It detaches structural features from the message passing of GNNs and substitutes them with heuristic-based pairwise features, which achieves better scalability but also compromises its capacity and empirical performance.

The emerging challenge in leveraging structural features to yield more expressive link representations lies in the trade-off between feature richness and their computation complexity. Structural link representations are edge-conditioned: materializing structure features with subgraph labeling for each target link is extremely inefficient and redundant, making it no longer feasible for large graphs. Directly using predefined pairwise features avoids explicit construction of structural features and the cost of rerunning GNNs for labeled subgraphs. However, such simplification disables the model from capturing some important structural signals on the graph, as these predefined features are hand-crafted to mimic certain heuristics, where structural information from node neighborhoods is squashed by some fixed score functions.

To tackle the above issues, we propose \proj, a compact representation of node neighborhood for constructing scalable structural features. It encodes the node neighborhood of arbitrary orders through hashing into a short-length bit array (termed ``signature"). By design, \proj can be preprocessed in parallel offline (only once) and be efficiently merged to recover edge-wise features online, including a variety of neighborhood overlap-based heuristics. This property avoids the expensive feature materialization of labeling tricks. Meanwhile, unlike hand-crafted pairwise features, it retains local structural information from original neighborhoods, enabling learning flexible data-driven heuristics and more expressive link representations. Combining GNNs with \proj strikes a balance between model expressiveness and computational complexity for learning structural link representations. Experimental results on five public benchmark datasets show that its prediction performance is comparable to or better than existing edge-wise GNN models, achieving 19-200 $\times$ speedup over the formerly state-of-the-art (SOTA) model SEAL~\cite{zhang2018link,zhang2021labeling} and 3.2$\times$ speedup over the fastest baseline BUDDY~\cite{chamberlain2022graph}.

Our main contributions can be summarized as follows:
\begin{itemize}[leftmargin=*]
    \item We identify key bottlenecks in applying structural features for learning expressive structural link representations and propose a decoupling mechanism to address the computational challenges of edge-wise feature construction while maintaining model expressiveness and empirical performance;
    \item A scalable hashing-based structural feature is proposed to augment the message-passing framework of GNNs: \proj is a compact encoding of node neighborhood that can be efficiently merged to recover edge-wise structural features for online training and inference.
    \item We provide error bounds for estimating neighborhood intersections from \proj and further show that any type of neighborhood overlap-based heuristics can be recovered from it with guaranteed accuracy via neural networks.
\end{itemize}

%% file: 2-Preliminary.tex
\section{Preliminaries and Related Work}
\paragraph{Notations} Let $\mathcal{G} = (\mathcal V, \mathcal E, X)$ be a undirected graph of $N$ nodes $\mathcal V = \{1, 2\dots, N\}$ and $E$ edges $\mathcal E \subseteq \mathcal V \times \mathcal V$ with node features of $X \in \mathbb R^{N\times F}$. Let $A \in \mathbb R^{N\times N}$ be the adjacency matrix of $\mathcal G$, where $A_{uv}=1$ if $(u,v)\in \mathcal E$ and $A_{uv}=0$ otherwise. The degree of node $u$ is $\deg(u):=\sum^{N}_{v=1}A_{uv}$. We denote the $k$-hop neighborhood of node $u$ as $\mathcal N^k(u)$, which is the set of all nodes that are connected to $u$ with the shortest path distance less than or equal to $k$. Note that $\mathcal N(u) = \mathcal N^1(u)$. The $k$th-order neighborhood intersection, union and difference between nodes $u$ and $v$ are given by $\mathcal N^k(u) \cap \mathcal N^k(v)$, $\mathcal N^k(u) \cup \mathcal N^k(v)$, and $\mathcal N^k(u) - \mathcal N^k(v)$. Let $\mathcal{H}: \mathcal V \to [n]$ be a pairwise independent hash function, i.e. $\Pr(\mathcal{H}(u) = h_1 \land \mathcal{H}(v) = h_2) = 1/n^2$ for $h_1, h_2 \in [n]$, $u \neq v$ and $u, v \in \mathcal V$.

\begin{table}
    \centering
    \caption{Unified formulation of structural features based on node neighborhoods in different orders.}
    \vspace{-3mm}
    \label{tab:undf}
    \resizebox{0.99\linewidth}{!}{
    \begin{tabular}{cclcc}
    \toprule
        Feature & Order of Neighbors & Set operator $\uplus$ & $f(x)$ & $d(\cdot)$\\ \hline
        CN & $k=1$ & $\mathcal N(u) \cap \mathcal N(v)$ & $x$ & 1\\
        RA & $k=1$ & $\mathcal N(u) \cap \mathcal N(v)$ & $x$ & $1/\deg(w)$\\
        AA & $k=1$ & $\mathcal N(u) \cap \mathcal N(v)$ & $x$ & $1/\log \deg(w)$\\
        Neo-GNN & $k_1,k_2\geq1$ & $\mathcal N^{k_1}(u) \cap \mathcal N^{k_2}(v)$ & MLP & $\deg(w)$\\
        BUDDY & $k_1,k_2\geq1$ & $\begin{aligned} \mathcal N^{k_1}(u) &\cap \mathcal N^{k_2}(v), \\ \mathcal N^{k_1,k_2}(u) &- \cup_{k'=1}^k \mathcal N^{k'}(v)\end{aligned}$ & $x$ & 1\\
        Labeling Trick & $k \geq 1$ & $\mathcal N^{k}(u) \cup \mathcal N^{k}(v)$ & MPNN & $\mathrm{dis}(w|(u,v))$\\
    \bottomrule
    \end{tabular}}
    \vspace{-5mm}
\end{table}

\paragraph{Link Prediction Heuristic} Common heuristics for link prediction is neighborhood overlap-based with varieties in score functions to measure the similarity. \textbf{1-hop neighborhood}: common neighbor (CN) $S_\mathrm{CN}(u, v) = \sum_{w \in \mathcal N(u) \cap \mathcal N(v)} 1$, resource allocation (RA) $S_\mathrm{RA}(u, v) = \sum_{w \in \mathcal N(u) \cap \mathcal N(v)} 1/{\mathrm{deg}(w)}$, and Adamic-Adar (AA) $S_\mathrm{AA}(u, v) = \sum_{w \in \mathcal N(u) \cap \mathcal N(v)} 1/{\log\mathrm{deg}(w)}$. \textbf{High-order neighborhood}: Neo-GNN~\cite{yun2021neo} utilizes a weighted feature of node degree in the intersection of $k$-many orders as $S_\mathrm{Neo}(u, v)=\sum_{k_1=1}^k\sum_{k_2=1}^k \big(\beta^{k_1+k_2-2}\cdot \sum_{w \in \mathcal N^{k_1}(u) \cap \mathcal N^{k_2}(v)}f(\deg(w))\big)$, where $\beta$ is a hyperparameter and $f(\cdot)$ is a learnable function. BUDDY~\cite{chamberlain2022graph} uses $k^2$ intersection and $2k$ difference features as $S_\mathrm{BUDDY}(u, v)=\{\sum_{w \in \mathcal N^{k_1}(u) \cap \mathcal N^{k_2}(v)}1|k_1,k_2 \in [k]\}||\{\sum_{w \in \mathcal N^{k_1,k_2}(u) - \cup_{k'=1}^k \mathcal N^{k'}(v)}1|k_1,k_2 \in [k]\}$, where $||$ denotes concatenation.

\paragraph{Message Passing Neural Network} Message passing neural network (MPNN) \cite{gilmer2017neural} is a generic framework of GNNs, which learns node representations via iterative aggregations of their local neighborhoods in the graph. To compute the message $\vm_v^{(l)}$ and the hidden state $\vh_v^{(l)}$ for each node $v \in \mathcal V$ at the $l$-th layer ($l=1,2,\dots,L)$:
\[\begin{aligned}
    \vm_v^{(l)} &= \texttt{AGG}\left(\left\{\phi^{(l)}\left(\vh_v^{(l-1)},\vh_w^{(l-1)}\right)|{w \in \mathcal{N}(v)}\right\}\right),\\
    \vh_v^{(l)} &= \sigma^{(l)}\left(\vh_v^{(l-1)}, \vm_v^{(l)}\right),
\end{aligned}\]
where $\phi, \sigma$ are learnable functions of message and update, respectively. $\texttt{AGG}$ is a local permutation-invariant aggregation function (e.g., sum, mean, max). The node representation is initialized as $\vh_v^{(0)}=X_v$. The final layer output gives the node representation $\vh_v$ (omits the layer superscript for simplicity). To predict the likelihood between two given nodes $u,v$ to form a link (noted as $\hat{A}_{ij}$) from node representations produced by MPNN: 
\[\hat{A}_{uv} = \mathrm{sigmoid}(\mathrm{MLP}(\vh_u \circ \vh_v)),\]
where $\circ$ is the decoder for link prediction, such as dot product, Hardamard product, or concatenation. A multilayer perceptron (MLP) is commonly applied as a link predictor.

\begin{figure}
    \centering
    \includegraphics[width=\linewidth]{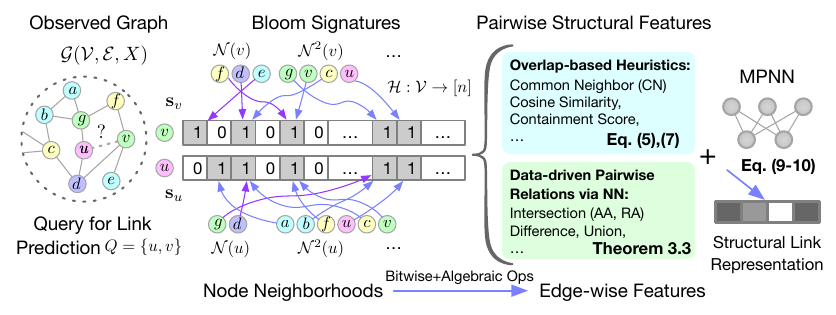}
    \vspace{-9mm}
    \caption{\small{\proj: a scalable hashing-based structural feature of node neighborhoods, which can be used to recover neighborhood overlap-based heuristics or capture data-driven pairwise relations for structural link representation learning.}}
    \label{fig:bsign}
    \vspace{-4mm}
\end{figure}

\begin{definition}[\citeauthor{srinivasan2019equivalence,li2020distance}] The \textit{most expressive structural representation of link} gives the same representation if and only if two links are isomorphic (symmetric, on the same orbit) in the graph. \label{def:mesr}
\end{definition}

\paragraph{Structural Features for Link Prediction} From Def. \ref{def:mesr}, a most expressive structural representation will distinguish all non-isomorphic links, where simply aggregating node representations generated from GNNs fails to do so. GNNs map isomorphic nodes into the same representation as its expressive power is bounded by the 1-WL test. As Fig. \ref{fig:expressive} shows, such behavior would lead to two remotely related nodes $u,w$ and two neighboring nodes $u,v$ sharing the same link representation, just because the tail nodes $w$ and $v$ have the isomorphic neighborhood. Structural features are introduced to break such symmetry in their neighborhoods and inject dependencies between queried nodes $(u,v)$ through proximity-based functions $\mathrm{dis}(w|(u,v)), \forall w\in \mathcal{N}^k(u) \cup \mathcal{N}^k(v)$ such as shortest path distance (SPD) or random walk landing probability (RW). Specifically, Double Radius Node Labeling (DRNL, \cite{zhang2018link}) uses $\mathrm{dis}(w|(u,v))=\mathrm{SPD}(u,w)+\mathrm{SPD}(v,w)$, and Relative Positional Encoding~\cite{yin2022algorithm,yin2023surel+} adopts $\mathrm{dis}(w|(u,v))=\{\mathrm{RW}(u,w)||\mathrm{RW}(v,w)\}$. It has been proven that a sufficiently expressive GNN combined with structural features of labeling tricks or link prediction heuristics can learn the most expressive structural representation of links \citep{wang2023improving}.

%% file: 3-Methodology.tex
\section{Bloom Signatures for Structural Link Representation Learning}
\subsection{Breakdown Pairwise Structural Features}
Using structural features for learning link representations is justified by theoretical foundation and empirical performance. However, successfully deployed structural features are edge-specific and conditioned on pairwise node relations, which must be computed for individual links during training and inference. This property makes it not easily parallelizable as the way node features are processed in GNNs. As a result, the challenge in learning expressive structural link representation is how to obtain high-quality pairwise structural features at low cost.

By observation, neighborhood overlap-based structural features, including most link prediction heuristics and high-order intersection- and difference-based features, share the format as
\begin{equation}
    S(u,v) = \sum_{w \in \mathcal{N}^k(u)\uplus \mathcal{N}^k(v)} f(d(w)).
    \label{eq:undf}
\end{equation}
We unify them by defining a function $d: \mathcal V \to \mathbb R^+$ and a learnable function $f(\cdot)$. Table \ref{tab:undf} summarizes this unified formulation for common heuristics and pairwise features in different orders. Note that, labeling trick~\cite{zhang2021labeling} implicitly follows Eq. (\ref{eq:undf})'s formulation by first labeling nodes within the union of $k$-hop neighborhoods of two target nodes via a predefined distance function $\mathrm{dis}(\cdot)$ and then feeding the labeled neighborhood (subgraph) for message passing.

From Eq. (\ref{eq:undf}), node neighborhood is essential to obtaining pairwise structural features, and set operations constitute the main computational cost. To label nodes, one common trick, DRNL requires two traversals over the neighborhood union (see Fig. \ref{fig:expressive}), with a complexity of $O(\bar{d}^k)$, where $\bar{d}$ is the average degree. While heuristic-like pairwise features are hand-crafted to obtain certain topological statistics (e.g., intersection, difference, union) from node neighborhoods. It eliminates explicit labeling and detaches structural features from message passing, and thus reduces the complexity. However, such simplification greatly compromises the model's ability and flexibility to capture any additional structural signals not included in the specific predefined statistics.

One solution is to decompose structural features into a carefully designed representation of node neighborhoods, from which pairwise features can be efficiently recovered. Such a design brings three benefits: (1) it retains the topological information of the original neighborhood while providing more richness and flexibility; (2) it decouples the dependence on edges, where node-wise neighborhood representations can be preprocessed offline; (3) there are low-cost substitutes of set operations for reconstructing pairwise features from such neighborhood representations. However, it is prohibitively expensive to store and operate on the complete neighborhood for each node, not to mention it is sparse and of varying sizes. An ideal neighborhood representation should be space efficient while being able to produce accurate estimations of pairwise features online with lightweight operations.

\subsection{Encode Node Neighborhoods via Hashing}
We propose \proj, a compact, merge-able encoding of node neighborhoods, inspired by Bloom filter~\cite{bloom_filter} that is a probabilistic data structure for set membership testing. \proj encodes a node's neighborhood into a short-length bit array (``signature") via hashing without explicitly storing the full members. The obtained signature contains provably sufficient information to estimate pairwise structural features such as intersection and cosine similarity between neighborhoods.

For node $u$, its complete neighborhood $\mathcal N(u)$ can be represented by a $N$-length binary array $\vu \in \{0, 1\}^N$, where $\vu[w] = \mathbbm 1\{w \in \mathcal N(u)\}$ (set the $w$-th bit to 1 if node $w$ exists in $\mathcal N(u)$, and 0 otherwise). Clearly, it is impractical to use such a long and sparse array to represent $\mathcal N(u)$. However, the sparsity of $\vu$ can be exploited to reduce its size while maintaining the property of recovering topological statistics and pairwise structural features from it. By using a hash function $\mathcal{H}$ to randomly map elements from $\{1,\dots,N\}$ to $\{1,\dots,n\}$ ($n \ll N$), we obtain the \textbf{\proj} $\vs_u \in \{0,1\}^n$ as a compact encoding of the node neighborhood $\mathcal N(u)$:
\begin{equation}
    \vs_u[j] = \bigvee_{i:\mathcal H(i)=j} \vu[i], \label{eq:bs}
\end{equation}
where $\bigvee$ denotes the bit-wise OR operator. $\vs_u $ refers to the signature of $\mathcal N(u)$, whose size is reduced to $n$ while approximately preserving membership information of node neighborhoods. Note that \proj is a special case of Bloom filters but with one hash function. Rather than for membership testing, \proj compresses node neighborhoods while maintaining sufficient structural information from which various pairwise features can be reconstructed and learned for link prediction. There are three benefits of using \proj:
\begin{itemize}[leftmargin=*]
    \item \textbf{Informative}: common pairwise heuristics such as neighborhood intersections can be recovered with a guarantee (Sec. 3.3);
    \item \textbf{Flexible}: data-driven pairwise features can be learned end-to-end by feeding Bloom signatures into neural networks (Sec. 3.4);
    \item \textbf{Expressive}: MPNN is provably more powerful by enhancing it with \proj (Sec. 3.5).
\end{itemize}

\subsection{Recover Pairwise Heuristics from Neighborhood Signatures \label{sec:pairwise}}
After hashing each node's neighborhood, one can obtain neighborhood overlap-based heuristics from a pair of Bloom signatures with guaranteed accuracy. This section describes how to calculate common neighbors, cosine similarity, and containment scores between two neighborhoods by simply merging their corresponding Bloom signatures. We first show how to recover the cardinality of node neighborhood $\mathcal N_u$ from a signature $\vs_u$ and then introduce the merge operation on $\vs_u, \vs_v$ for estimating pairwise features. 

Since the hash function $\mathcal{H}$ in Eq. (\ref{eq:bs}) is uniformly random, we can establish the relation between the number of non-zeros in $\vs_u$ and the cardinality of $\mathcal N_u$ by $\mathbb E\left(|\vs_u|/n\right) = 1-(1-1/n)^{|\mathcal N(u)|}$, where $\vs_u$ has a size of $n$ with $|\vs_u|$-many non-zero bits. Thus, the size of node $u$'s neighborhood can be estimated from the signature $\vs_u$ as
\begin{equation}
    |\mathcal N(u)| \approx \hat n_u = \ln \left(1-\frac{|\vs_u|}{n}\right)\big/\ln \rho,~\rho=1-\frac 1 n \in (0,1).
    \label{eq:card}
\end{equation}
The error bound of the above cardinality estimation is given by Lemma \ref{lm:card}, and its full proof is provided in Appx. \ref{appx:eb}.
\begin{lemma}\label{lm:card}
With probability at least $1-\delta$, it holds that
\begin{equation}
\big||\mathcal{N}(u)| - \hat n_u\big| < \sqrt{4m\log\frac{2}{\delta}},
\end{equation}
where $m$ is the sparsity of the binary vector $\vu$ representing $\mathcal{N} (u)$.
\end{lemma}

Given a pair of $n$-size signatures $\vs_u,\vs_v$, their inner product preserves the cardinality and the intersection size of original node neighborhoods after random mapping~\cite{network_bloom}: 
\begin{equation*}
    \resizebox{\hsize}{!}{$\mathbb E\left(\frac{\langle \vs_u, \vs_v\rangle}{n}\right) = 1-\rho^{|\mathcal N(u)|}-\rho^{|\mathcal N(v)|}+\rho^{|\mathcal N(u)|+|\mathcal N(v)|-|\mathcal N(u)\cap \mathcal N(v)|}.$}
\end{equation*}
This allows us to express the intersection size of two neighborhoods $|\mathcal N (u) \cap \mathcal N (v)|$ (equivalently, common neighbors) in terms of the inner product of two Bloom signatures $\vs_u,\vs_v$ as
\begin{equation}
    |\mathcal N (u) \cap \mathcal N (v)| \approx \hat S_{\mathrm{CN}}(u,v) = \hat n_u + \hat n_v - \frac{\ln \left(\rho^{\hat n_u}+\rho^{\hat n_v}+\frac{\langle \vs_u, \vs_v \rangle}{n}-1\right)}{\ln \rho},
    \label{eq:int}
\end{equation}
where $\hat n_u , \hat n_v$ are obtained from Eq. (\ref{eq:card}) and $\rho = 1-1/n$. We further provide the quality analysis of the above estimation of intersection size and give the error bound in Theorem \ref{thm:intersec} below (detailed proof in Appx. \ref{appx:eb}).

\begin{theorem}\label{thm:intersec}
With probability at least $1-3\delta$, it holds that 
\begin{equation}
    \left| |\mathcal{N}(u) \cap \mathcal{N}(v)| - \hat S_{\mathrm{CN}}(u,v)\right|<6\sqrt{m}+7\sqrt{2m \ln \frac{2}{\delta}},
\end{equation}
where the $N$-dim binary vector representing $\mathcal{N}(u)$ or $\mathcal{N}(v)$ has the sparsity at most $m$ with probability at least $1-\delta'$.
\end{theorem}

The inner product of two Bloom signatures can be efficiently obtained by ``merging'' them in parallel: bit-wise AND operations followed by a summation. From Eqs. (\ref{eq:card}) and (\ref{eq:int}), we can recover the neighborhood cardinality and their pairwise intersections with guaranteed accuracy, which form the basis of all neighborhood overlap-based heuristics. Next, we show how to generalize beyond intersections by including the neighborhood size. Use cosine similarity $S_{\mathrm{cos}}(u, v) =|\mathcal N(u) \cap \mathcal N(v)| / \sqrt{|\mathcal N(u)||\mathcal N(v)|}$ and containment score $S_{\mathrm{cont}}(u, v) = |\mathcal N(u) \cap \mathcal N(v)|/ |\mathcal N(u)|$ as examples, we can estimate them from $\hat{n}_u,\hat{n}_v$ and the $\hat{S}_{\mathrm{CN}}$ obtained earlier: 
\begin{equation}
    \hat{S}_{\mathrm{cos}}(u, v) = \frac{\hat S_{\mathrm{CN}}(u, v)}{\sqrt{\hat n_u \hat n_v}},\qquad \hat{S}_{\mathrm{cont}}(u, v) = \frac{\hat S_{\mathrm{CN}}(u, v)}{\hat n_u}.
\end{equation}
Note that cosine similarity is symmetric, but containment score is not, and both of them provide finer details of original node neighborhoods. In addition, the compactness of \proj enables the encoding of high-order (i.e., multi-hop) neighborhoods that are known to contain useful features for link prediction~\cite{yun2021neo,yin2022algorithm}.

\begin{table*}[ht]
    \centering
    \caption{Complexity comparison. $h_s,h_b$: the complexity of hash operations in Subgraph Sketch and \proj, respectively. $F$: the dimension of node embedding. $k$ is the number of hops for extracted subgraphs ($k$-hop node neighborhoods). $h$ is the complexity of hashing-based methods to obtain pairwise structural features. When predicting $q$-many target links, the time complexity mainly comes from message passing and link predictor. \label{tab:complexity}}
    \vspace{-3mm}
    \begin{tabular}{lccccc}
    \toprule
    Complexity & GNN & SEAL & Neo-GNN & BUDDY & Ours\\
    \midrule
    Preprocessing & $1$ & $1$ & $1$ & $k|E|(\bar{d}+h_s)$ & $kNh_b$\\
    Message Passing & $N\bar{d}F+NF^2$ & 0 & $N\bar{d}F+NF^2$ & $N\bar{d}F$ & $N\bar{d}F+NF^2$\\
    Link Predictor & $qF^2$ & $q(\bar{d}^{k+1}F+\bar{d}^kF^2)$ & $q(\bar{d}^{k}+F^2)$ & $q(h+F^2)$ & $q(h+F^2)$ \\
    \bottomrule
\end{tabular}
\vspace{-3mm}
\end{table*}

\subsection{Learning Data-driven Pairwise Relations}
Common link prediction heuristics can be recovered by simply merging a pair of Bloom signatures, but most of them are hand-crafted and thus inflexible. The summation in Eq. (\ref{eq:undf}) reduces the structural information of node neighborhoods to certain statistical values, making it hardly learnable. Ideally, we would like to capture pairwise relations in an end-to-end manner. For example, SEAL~\cite{zhang2018link,zhang2021labeling} directly modifies the input graph and attaches link-specific distance features for message passing. However, the coupling between node labeling and target links makes its deployment prohibitively expensive. Although some simplifications~\cite{yun2021neo,kong2022geodesic,yin2022algorithm,yin2023surel+} of labeling tricks have been proposed recently, they still suffer from the high complexity of the set operation $\uplus$ used to connect neighborhoods of two target nodes.

\proj carries sufficient structural information of node neighborhoods in its compact encoding but does not involve any set operations to extract pairwise relations. This property further enables us to use neural networks to capture important signals in encoded neighborhoods and identify data-drive pairwise heuristics from Bloom signatures at a much lower cost. Theorem \ref{thm:approx} shows that neighborhood overlap-based heuristics can be recovered to arbitrary precision by a neural network (such as MLP) taking a pair of Bloom signatures as input (full proof provided in Appx. \ref{appx:approx}).

\begin{theorem}\label{thm:approx}
Suppose $S(u,v)$ is a neighborhood intersection-based heuristic defined in Eq. (\ref{eq:undf}) with the maximum value of $d(\cdot)$ as $d_\mathrm{max} = \max_{w \in \mathcal V} d(w)$. Let $p_u = (1 - 1/n)^{|\mathcal N(u)|}$ denote the false positive rate of set membership testing in the Bloom signature $\vs_u$ (similarly to $p_v$). Then, there exists an MLP with one hidden layer of width $N$ and ReLU activation, which takes the Bloom signatures of $u$ and $v$ as input and outputs $\hat{S}(u, v) = \mathrm{MLP}(\vs_u, \vs_v)$, such that with probability 1, $\hat{S}(u, v) - S(u, v) \ge 0$; with probability at least $1 - 3\delta$, it holds that
\begin{equation}\label{eq:nie}
\begin{aligned}
    &\hat{S}(u, v) - S(u, v) \le \\
    &\resizebox{\hsize}{!}{$d_\mathrm{max}\Bigg(\bigg(1 + \sqrt{\frac{-3\log \delta}{|\mathcal S_C|p_up_v}}\bigg)\big|\mathcal S_C\big|p_up_v + \left(1 + \sqrt{\frac{-3\log \delta}{|\mathcal S_{D_v}|p_v}}\right)\big|\mathcal S_{D_v}\big|p_v + \left(1 + \sqrt{\frac{-3\log \delta}{|\mathcal S_{D_u}|p_u}}\right)\big|\mathcal S_{D_u}\big|p_u\Bigg)$,}
\end{aligned} 
\end{equation}
where $\mathcal S_C = \mathcal V \backslash (\mathcal N(u) \cup \mathcal N(v))$, $\mathcal S_{D_v}=\mathcal N(u) \backslash \mathcal N(v)$, and $\mathcal S_{D_u}=\mathcal N(v) \backslash \mathcal N(u)$.
\end{theorem}
In fact, the above results can be generalized to any pairwise relations of node neighborhoods following the form of Eq. (\ref{eq:undf}), including set difference and union with theorems given in Appx. \ref{appx:approx}. The approximation error in Eq. (\ref{eq:nie}) is proportional to the false positive rate of \proj, meaning any neighborhood intersection-based heuristic can recover exactly if $p_u=0$. For commonly used heuristics such as CN, AA, and RA, we have $d_\mathrm{max} = 1$.

\paragraph{Other hashing-based methods} Subgraph sketch~\cite{chamberlain2022graph} utilizes a combination of MinHash and HyperLogLog to estimate the intersection and complement of high-order node neighborhoods. These handcrafted pairwise features are used as input to MPNN or MLP for link prediction, obtaining comparable results to labeling tricks. However, subgraph sketch either faces high storage and computational overhead (coupled with message passing) or has limited expressiveness (decoupled, using MLP only). The two hashing algorithms employed require hundreds of repetitions and a large memory budget to achieve reasonable estimation quality for pairwise features (see Fig. \ref{fig:random_graph_error} in Sec. \ref{exp:estimation} for the study of estimation error). Furthermore, sketches of subgraphs can only be used to estimate neighborhood intersections and complements, which is inflexible and greatly limits the richness of structural information. In comparison, \proj captures node membership of encoded neighborhoods, enabling more powerful models to recover any type of neighborhood overlap-based heuristics (such as RA and AA) or to learn data-driven pairwise relations, both of which are unachievable by subgraph sketch.

\subsection{Boost Message Passing with Scalable Pairwise Structural Features}
Putting all the pieces together, we propose a new scalable MPNN framework enhanced by \proj depicted in Fig. \ref{fig:bsign}, which achieves a trade-off between complexity and expressiveness for learning structural link representations. It inherits the rich capacity of pairwise structural features while decoupling their construction from specific edges, enabling efficient online training and inference. In particular, we can add \proj as edge features/weights to moderate the message function and/or as additional pairwise features for the link predictor:

\begin{align}
    \label{eq:ampnn}
    \vh_v^{(l)} &= \sigma^{(l)}\left(\vh_v^{(l-1)}, \texttt{AGG} \left(\left\{ \phi^{(l)}\left(\vh_w^{(l-1)}, \vh_v^{(l-1)}, \ve_{w,v}\right)|w\in \mathcal{N}(v)\right\}\right)\right), \\
    \label{eq:sflp}
    \hat{A}_{uv} &= \mathrm{sigmoid}\left(\mathrm{MLP}\left(\vh_u \circ \vh_v || \hat{S}(u,v)\right)\right)
\end{align}
where $\ve_{w,v} = \psi_f(\{\vs_w || \vs_v\})$ or $\psi_g(||\vs_w - \vs_v||)$, and $\psi_f,\psi_g$ are neural encoders such as MLP. $\hat{S}(u,v)$ can be any estimated heuristics or learnable data-driven pairwise relations.

\paragraph{Expressiveness of MPNN with \proj} By introducing \proj, the new framework acts as a more powerful and flexible feature extractor for pairwise relations while enjoying the tractable computational complexity of MPNN. As a result, it is strictly more expressive than vanilla GNNs: (1) can measure the overlap between node neighborhoods via recovering intersection-based heuristics from Bloom signatures; (2) can avoid node ambiguity via regulated message passing with Bloom signatures and recovered pairwise structural features. We provide the analysis of expressive power for the proposed model in Appx. \ref{sec:expresiveness}.

\paragraph{Complexity Comparison} Table \ref{tab:complexity} summarizes the computation complexity of vanilla GNN and existing edge-wise models for link prediction. Both BUDDY~\cite{chamberlain2022graph} and \proj are hashing-based methods, which require preprocessing to obtain $k$-hop subgraph sketches in $\mathcal{O}(k|E|h_s)$ and signatures of $k$-hop node neighborhoods in $\mathcal{O}(kNh_b)$. Note that, hashing operations in BUDDY require hundreds of samples and large storage to generate reasonable estimates, while \proj only needs bit-wise and simple algebraic operations with guaranteed accuracy. Vanilla GNN, Neo-GNN \cite{yun2021neo}, and MPNN with \proj share the same complexity for message passing. BUDDY precomputes the node feature propagation with $\mathcal{O}(k|E|\bar{d})$ and takes another $\mathcal{O}(N\bar{d}F)$ to obtain node embeddings. For predicting $q$-many target links, vanilla GNN only needs to call the link predictor with the cost of $\mathcal{O}(qF^2)$. SEAL \cite{zhang2018link,zhang2021labeling} runs MPNN for each link-induced $k$-hop subgraph of size $\mathcal{O}(\bar{d}^k)$, resulting in the total complexity of $\mathcal{O}(q(\bar{d}^{k+1}+\bar{d}^kF^2))$. Neo-GNN needs an additional $\mathcal{O}(\bar{d}^k)$ to compute pairwise features before feeding node embeddings into the link predictor. Both BUDDY and \proj require additional operations on preprocessed node-wise features with a complexity of $h$ to obtain pairwise structural features. Here, BUDDY tries to amortize the cost of pairwise features in preprocessing, which is infeasible in practice when target links are unknown.

%% file: 4-Experiment.tex
\begin{table}
\centering
\caption{Summary statistics for evaluation datasets.}
\label{tab:data}
\vspace{-3mm}
  \resizebox{0.48\textwidth}{!}{
  \begin{tabular}{llrrrc}
    \toprule
    \textbf{Dataset}&\textbf{Type}&\textbf{\#Nodes}&\textbf{\#Edges} & $\bar{d}$ &\textbf{Split(\%)}\\
    \midrule
    \texttt{citation2} & Homo./Social. & 2,927,963 & 30,561,187 & 10.44 &  98/1/1\\
    \texttt{collab} & Homo./Social. & 235,868 & 1,285,465 & 5.45 & 92/4/4\\ \hline
    \texttt{ddi} & Homo./Drug & 4,267 &  1,334,889 & 312.84 & 80/10/10\\
    \texttt{ppa} & Homo./Protein & 576,289 &  30,326,273 & 52.62 & 70/20/10\\
    \texttt{vessel} & Homo./Vesicular & 3,538,495 & 5,345,897 & 1.51 & 80/10/10\\
  \bottomrule 
\end{tabular}}
\vspace{-5mm}
\end{table}

\begin{table*}[htp]
\centering
\caption{Results (\%) on OGB datasets for link prediction. Highlighted and underlined are the results ranked \textbf{FIRST}, \underline{second}.\label{table:ogb}}
 \resizebox{0.98\textwidth}{!}{
 \vspace{-3mm}
\begin{tabular}{cc|cc|cc|cc|c|c}
\toprule
 &\multirow{2}{*}{Models} & \multicolumn{2}{c|}{\texttt{collab}} & \multicolumn{2}{c|}{\texttt{ddi}} & \multicolumn{2}{c|}{\texttt{ppa}} & \texttt{citation2} & \texttt{vessel} \\ \cmidrule(r{0.5em}){3-10}
 & & Hits@50 & AUC & Hits@20 & AUC & Hits@100 & AUC & MRR & AUC \\
  \midrule
  \multirow{3}{*}{\rotatebox{90}{\small Heuristic}}  
& CN &61.37 &82.78 &17.73 &95.20 &27.65 &97.22 &50.31 &48.49\\
& AA &64.17	&82.78 &18.61 &95.43 &32.45 &97.23 &51.69 &48.49\\
& RA &63.81	&82.78 &6.23  &96.51 &49.33 &97.24 &51.65 &48.49 \\ \midrule
\multirow{3}{*}[0.6ex]{\rotatebox{90}{\small Embedding}}  
&MF &41.81 ± 1.67 &83.75 ± 1.77 &23.50 ± 5.35 &99.46 ± 0.10 &28.40 ± 4.62 &99.46 ± 0.10 &50.57 ± 12.14 & 49.97 ± 0.05 \\
&MLP &35.81 ± 1.08 &95.01 ± 0.08 &N/A &N/A &0.45 ± 0.04 &90.23 ± 0.00 &38.07 ± 0.09 &50.28 ± 0.00 \\
&Node2Vec &49.06 ± 1.04	&96.24 ± 0.15 &34.69 ± 2.90 &99.78 ± 0.04 &26.24 ± 0.96 &99.77 ± 0.00 &45.04 ± 0.10 & 47.94 ± 1.33\\ \midrule
\multirow{4}{*}[1.2ex]{\rotatebox{90}{Vanilla}}  
&GCN	&54.96 ± 3.18 &97.89 ± 0.06 &\underline{49.90 ± 7.23} &99.86 ± 0.03 &29.57 ± 2.90 &99.84 ± 0.03 &84.85 ± 0.07 & 43.53 ± 9.61\\
&SAGE	&59.44 ± 1.37 &98.08 ± 0.03 &49.84 ± 15.56 &99.96 ± 0.00 &41.02 ± 1.94 &99.82 ± 0.00 &83.06 ± 0.09 & 49.89 ± 6.78\\
&GAT	&55.00 ± 3.28 &97.11 ± 0.09 &31.88 ± 8.83 &99.63 ± 0.21 &OOM &OOM &OOM & OOM\\ \midrule
\multirow{7}{*}[2ex]{\rotatebox{90}{Edge-wise}}  
&SEAL & 63.37 ± 0.69 &95.65 ± 0.29 &25.25 ± 3.90 &97.97 ± 0.19 &\underline{48.80 ± 5.61} &99.79 ± 0.02 & 86.93 ± 0.43 & \textbf{80.50 ± 0.21}\\
&Neo-GNN & \textbf{66.13 ± 0.61} &98.23 ± 0.05 &20.95 ± 6.03 &98.06 ± 2.00 &48.45 ± 1.01 &97.30 ± 0.14 & 83.54 ± 0.32 & OOM\\
&GDGNN & 54.74 ± 0.48 & 91.65 ± 0.32 &21.01 ± 2.09 & 93.90 ± 0.90 &45.92 ± 2.14 & 98.84 ± 0.43 & 86.96 ± 0.28 & 75.84 ± 0.08\\
&BUDDY & 64.59 ± 0.46 &96.52 ± 0.40 &29.60 ± 4.75 &99.81 ± 0.02 &47.33 ± 1.96 &99.56 ± 0.02 & \textbf{87.86 ± 0.18} & 65.30 ± 0.09 \\
&Ours & \underline{65.65 ± 0.32} & 96.91 ± 0.43 & \textbf{55.35 ± 9.04}& 99.96 ± 0.00 & \textbf{49.80 ± 1.74} & 99.82 ± 0.01 &\underline{87.29 ± 0.20} & \underline{79.83 ± 0.52} \\ \bottomrule
\end{tabular}}
\vspace{-3mm}
\end{table*}

\section{Evaluation}
In this section, we aim to evaluate the following questions:
\begin{itemize}[leftmargin=*]
    \item How scalable is \proj compared to SOTA link prediction models, including hashing-based and other simplified methods of labeling tricks?
    \item Can MPNN with \proj provide prediction performance comparable to existing edge-wise baselines?
    \item How is the estimation quality and efficiency of \proj?
\end{itemize}

\subsection{Experiment Setup}
\textbf{Datasets} Table \ref{tab:data} summarizes the statistics of datasets used to benchmark different models for link prediction. Five homogeneous networks are selected from the Open Graph Benchmark (OGB) \cite{hu2020open} at different scales (4K $\sim$ 3.5M nodes and 1.2M $\sim$ 30.6M edges), various densities (average degree $\bar{d}$ from 1.51 to 312.84), and with distinct characteristics: social networks of citation - \texttt{citation2} and collaboration - \texttt{collab}; biological network of protein interactions - \texttt{ppa}, drug interactions - \texttt{ddi} and brain vessels - \texttt{vessel}. Social networks play a key role in network analysis of modeling real-world dynamics. Recently, biological networks have emerged as a new data source for network science research. Particularly, understanding the interactions between proteins, drugs, and the network structure of brain vessels is of unique significance for scientific discovery~\cite{jumper2021highly}, which can be used for new drug discovery and early neurological disease detection.

\textbf{Baselines} We consider both classic approaches and SOTA GNN-based models. 
\textbf{Link Prediction Heuristics}: Common Neighbors (CN)~\cite{barabasi1999emergence}, Adamic-Adar (AA)~\cite{adamic2003friends}, and Resource Allocation (RA)~\cite{zhou2009predicting}. 
\textbf{Embedding Methods}: Matrix Factorization (MF)~\cite{qiu2018network}, Multi-layer Perceptron (MLP) and Node2Vec~\cite{grover2016node2vec}. 
\textbf{Vanilla GNNs}: Graph Convolutional Network (GCN)~\cite{kipf2016semi}, GraphSAGE~\cite{hamilton2017inductive} and Graph Attention Network (GAT)~\cite{velivckovic2017graph}.
\textbf{Edge-wise GNNs}: SEAL~\cite{zhang2018link,zhang2021labeling}, Neo-GNN~\cite{yun2021neo}, GDGNN~\cite{kong2022geodesic}, BUDDY~\cite{chamberlain2022graph}.

\textbf{Settings} Data split of OGB is used to isolate validation and test links from the input graph. We adopt official implementations of all baselines with tuned hyperparameters and match the reported results in \cite{li2023evaluating}. All experiments are run 10 times independently, and we report the mean performance and standard deviation.

\textbf{Evaluation Metrics} Ranking-based metrics (i.e., mean reciprocal rank (MRR) and Hits@$K$, $K\in \{20,50,100\}$) and the area under the curve (AUC) are used by default in OGB for evaluation. 

\textbf{Environment} We use a server with two AMD EPYC 7543 CPUs, 512GB DRAM, and two NVIDIA A100 (80GB) GPUs (only one GPU per model). The codebase is built on PyTorch 1.12, PyG 2.3, DGL 1.0.2, and Numba 0.56.

\subsection{Prediction Performance Analysis}
Table \ref{table:ogb} shows the prediction performance of different methods. On all five benchmarks, edge-wise models significantly outperform both vanilla GNNs and embedding-based models, especially for two challenging biological networks \texttt{ppa} and \texttt{vessel}. Link prediction in biological networks relies on pairwise structural information that vanilla GNNs have limited expressive power to capture. Among edge-wise models, MPNN with \proj achieves comparable or better performance than formerly SOTA SEAL, hashing-based BUDDY, and other simplified labeling tricks of Neo-GNN and GDGNN, which validates the effectiveness of the proposed compact neighborhood encoding. Unlike BUDDY, which explicitly relies on manually injecting features of CN and RA, our model can capture richer and more complex pairwise structural relations from Bloom signatures, which results in performance exceeding common link prediction heuristics on all five datasets.

\begin{figure*}[htp]
  \centering
  \includegraphics[width=\textwidth]{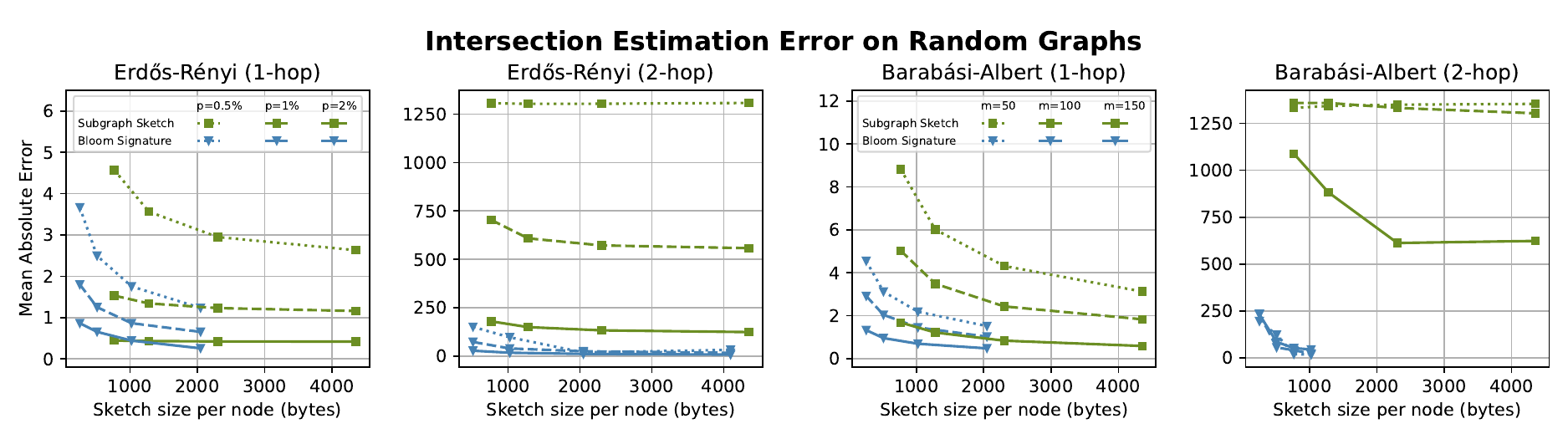}
  \includegraphics[width=\textwidth]{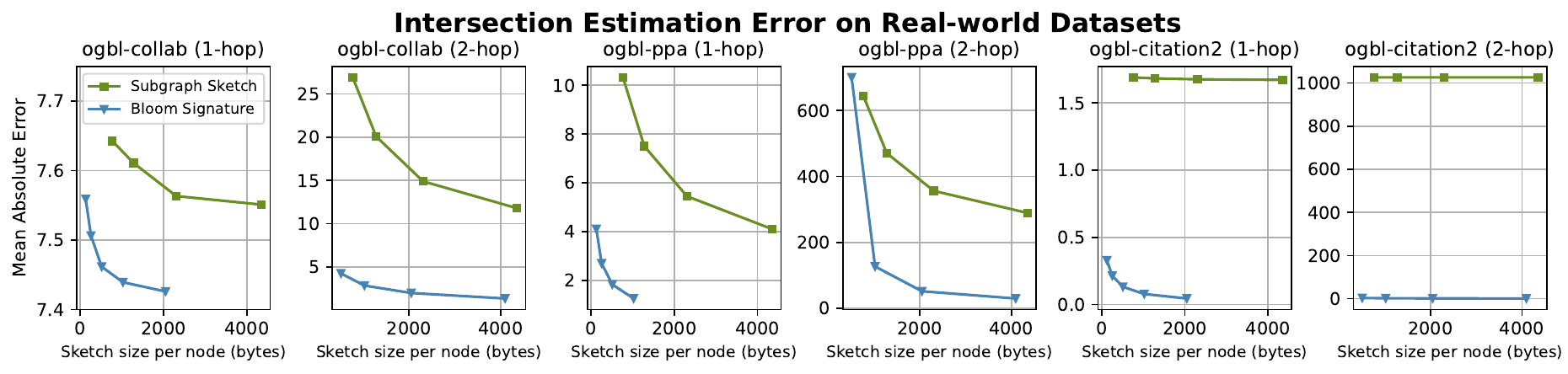}
  \vspace{-8mm}
  \caption{\small Mean absolute errors (MAE) of cardinality estimation for 1-hop and 2-hop neighborhood intersection on random networks and real-world networks by \proj and Subgraph Sketch with varying memory budget. Under the same memory budget, \proj produces estimations with up to $2.83\times$ and $68.52\times$ lower error than Subgraph Sketch for 1-hop and 2-hop neighborhood intersection on random graphs, and with up to $12.84\times$ and $303.10\times$ lower error on real-world graphs.}
  \label{fig:random_graph_error}
  \vspace{-5mm}
\end{figure*}

\subsection{Quality Analysis of Hashing-based Estimation for Pairwise Heuristics \label{exp:estimation}}
To measure the quality of hashing-based methods, \proj and Subgraph Sketch are compared on a variety of randomly generated graphs and real-world networks for estimating pairwise structural features. The key metric is the trade-off between estimation quality and the memory cost of hashing. We chose the cardinality of 1-hop and 2-hop neighborhood intersections as the estimated statistic since it is essential for many commonly used heuristics and pairwise structural features discussed earlier.

\paragraph{Random Graphs} Two types of representative random network models are picked: Erdős–Rényi~\cite{erdo_renyi} and Barabási–Albert~\cite{barabasi1999emergence}. Each random graph is generated with 10,000 nodes. The Erdős–Rényi model draws each edge from a binomial distribution with probability $p$. In Barabási–Albert, each node is added incrementally, with $m$ new edges attached to existing nodes with preferential attachment, meaning the sampling probability is proportional to the node degree. Graphs drawn from the Barabási–Albert model have power-law distributed node degrees, aligning with structural properties of real-world networks such as the World Wide Web, social networks, and academic graphs. 

Fig. \ref{fig:random_graph_error} (UP) presents the mean absolute error (MAE) of intersection estimation by two hashing-based methods on Erdős–Rényi ($p \in \{0.5\%, 1\%, 2\%\}$) and Barabási–Albert ($m \in \{50, 100, 150\}$) models with varying density and degree distribution. Under the same memory budget of hashing, \proj achieves lower estimation error than Subgraph Sketch in almost all cases. As the number of hops increases, the estimation quality of Subgraph Sketch degrades significantly. For the 2-hop neighborhood intersection, the estimation error of Subgraph Sketch levels off quickly and does not approach zero even with an increasing memory budget. On the other hand, \proj is capable of achieving low estimation error for high-order neighborhoods, and the error goes down to zero as more memory budget is allocated. Under the same memory budget, \proj achieves up to $2.83\times$ lower MAE than Subgraph Sketch for the 1-hop neighborhood intersection estimation and with up to $68.52\times$ lower error for the 2-hop estimation.

\paragraph{Real-world Networks} Fig. \ref{fig:random_graph_error} (DOWN) shows the quality comparison of intersection estimations on three real-world networks. \proj consistently produces more accurate results thanks to the design of compact encoding and non-sampling-based estimation from Eqs. (\ref{eq:card}), (\ref{eq:int}). Under the same memory budget, \proj produces up to $12.84\times$ and $303.10\times$ lower error than Subgraph Sketch for estimating 1-hop and 2-hop neighborhood intersections, respectively. Note that the estimation quality is dependent upon the degree distribution of the graphs and the actual intersection size, as larger neighborhoods are more difficult to encode, and more collisions occur during hashing. Aligning with Theorem \ref{thm:intersec}, \proj performs uniformly better than Subgraph Sketch on all real-world networks under different memory budgets. This observation is also consistent with the performance difference between BUDDY and our model in Table \ref{table:ogb}, when estimated pairwise structural features are used for prediction.

\begin{figure}[tp]
  \centering
  \includegraphics[width=\linewidth]{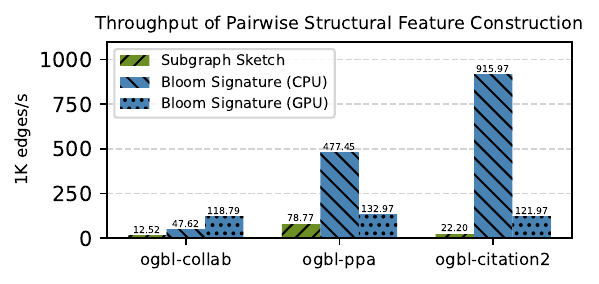}
  \vspace{-9mm}
  \caption{\small Throughput of pairwise structural features by Subgraph Sketch and \proj on three OGB datasets. \proj achieves up to $41.25\times$ and $9.48\times$ higher throughput than Subgraph Sketch on CPU (same number of threads) and GPU, respectively.}
  \label{fig:throughput}
  \vspace{-5mm}
\end{figure}

\begin{figure}[t]
  \centering
  \includegraphics[width=\linewidth]{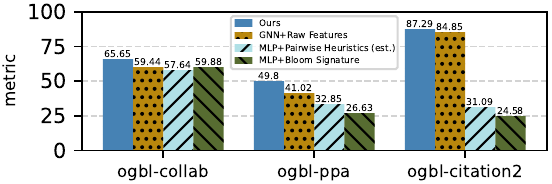}
  \vspace{-9mm}
  \caption{\small Comparison of different model configurations of MPNN with \proj on three OGB datasets.}
  \label{fig:ablation}
  \vspace{-6mm}
\end{figure}

\subsection{Efficiency and Scalability Analysis}
Fig. \ref{fig:throughput} compares the efficiency of two hashing-based methods by measuring the throughput of computing 2-hop pairwise structural features. Subgraph Sketch is currently only deployable to CPU, while \proj can perform accelerated calculations on CPU and GPU. \proj has $41.25\times$ and $9.48\times$ higher throughput than Subgraph Sketch on CPU and GPU, respectively. \proj uses only lightweight bitwise and algebraic operations, which favors multithreading on CPUs with lower I/O overhead. Running on GPU introduces I/O inefficiencies, especially on large-scale graphs (e.g., \texttt{citation2}). This hurts hardware-accelerated performance and results in \proj having relatively lower GPU throughput than CPU.

Table \ref{tab:scale} lists the end-to-end runtime of edge-wise GNN models on four large graphs. We list GCN as a reference point. The dynamic mode of SEAL (online subgraph extraction) is used due to its intractable resource needs for offline extraction on large graphs. Similarly, only the numbers of BUDDY are reported as its message passing version ELPH~\cite{chamberlain2022graph} is infeasible on \texttt{ppa} and \texttt{citation2}. The results of Neo-GNN are omitted as it is even slower than SEAL. Neo-GNN only runs MPNN once but needs $k$-order pairwise features. GDGNN requires first finding the geodesic path between node pairs, which constitutes the main computational cost for training and inference. BUDDY precomputes the node feature propagation together with sketches, which is reflected in the high cost of preprocessing and low training latency (no message passing involved). Similarly, \proj only needs to be computed once for $k$-hop node neighborhoods, but runs MPNN for obtaining node embeddings. In summary, MPNN with \proj does not suffer from the high complexity of pairwise structural features and is orders of magnitude faster than existing edge-wise models of SEAL, GDGNN, and Neo-GNN. In particular, \proj is 18.5–199.6 $\times$ faster than SEAL in dynamic mode on inference across four datasets. Under fair comparison on preprocessing of hashing and full online inference with pairwise structural features, it is $3.2\times$ faster than BUDDY on \texttt{citation2}.

\begin{table}[tp]
    \centering
    \caption{Runtime comparison. The row of Train records the time per 10K edges, and Inference is the full test set.}
    \label{tab:scale}
    \vspace{-2mm}
    \resizebox{0.48\textwidth}{!}{
    \begin{tabular}{c|c|cccccc}
    \toprule
    dataset  & time (s) & SEAL & GDGNN & BUDDY & GCN & Ours\\
    \midrule
    \multirow{3}{*}[0.5ex]{\rotatebox{90}{\small \texttt{citation2}}}  & Prep. & 0 & 338 & 721 & 17 & 92 \\
      & Train & 3.52 & 2.26 & 0.11 & 0.13 & 0.20\\
      & Inf. & 24,626 & 5,460 & 232 & 15 & 204 \\
    \midrule
    \multirow{3}{*}{\rotatebox{90}{\small \texttt{vessel}}}  & Prep. & 0 & 25 & 143 & 5 & 13\\
      & Train & 10.69 & 0.85 & 0.10 & 0.076 & 0.05\\
      & Inf. & 998 & 84 & 7 & 0.3 & 5\\
    \midrule
    \multirow{3}{*}{\rotatebox{90}{\small \texttt{ppa}}}  & Prep. & 0 & 127 & 368 & 2 & 109\\
      & Train & 10.57 & 1.77 & 0.027 & 0.026 & 0.07\\
      & Inf. & 3,988 & 902 & 25 & 1.2 & 35\\
    \midrule
    \multirow{3}{*}{\rotatebox{90}{\small \texttt{collab}}}  & Prep. & 0 & 14 & 10 & 2 & 12\\
      & Train & 4.05 & 0.74 & 0.049 & 0.005 & 0.06\\
      & Inf. & 37 & 15 & 1.5 & 0.05 & 2\\
    \bottomrule
    \end{tabular}}
    \vspace{-3mm}
\end{table}

\subsection{Ablation Study}
To validate the design of \proj, we conduct an ablation analysis and report results on three OGB datasets in Fig. \ref{fig:ablation} with the following configurations: (1) vanilla GNN with raw node features only; (2) MLP with multi-hop estimated pairwise structural features from Bloom Signatures; and (3) MLP with multi-hop raw pairwise Bloom Signatures as input. Neither MPNN with raw node features nor MLP with estimated/raw pairwise structural features alone can achieve the best performance. Note that, MLP taking a pair of raw Bloom Signatures as input achieves comparable or better performance than MLP using estimated pairwise structural features, suggesting that \proj encodes sufficient structural information of node neighborhood for the model to capture beyond handcrafted heuristics.

%% file: 5-Conclusion.tex
\section{Conclusion}
This work presents a new MPNN framework with scalable structural features for learning expressive representations of links, which is based on the observation and analysis of key bottlenecks in classic edge-wise GNN models. A novel structural feature termed \proj is proposed to encode node neighborhoods through hashing, which can be used to reconstruct any overlap-based pairwise heuristics with guaranteed accuracy. MPNN with \proj is provably more expressive than vanilla GNNs and also more scalable than existing edge-wise models. In practice, it achieves much better time and space complexity and superior prediction performance on a range of standard link prediction benchmarks.

%% file: 6-Appendix.tex
\section{Additional Experiment Details}
\paragraph{Model Architecture} We use 3 layers of GCN or GraphSAGE (the one that performs the best) as the backbone of MPNN in Eq. (\ref{eq:ampnn}) with support of edge features. 3-layer MLP of hidden dimension 256 with layer normalization is used as the link predictor (dot/diff) and pairwise structural feature encoder. Bloom signatures of 2-hop node neighborhoods are used for all five datasets, with sizes of $\{2048,4096,8192\}$.

\paragraph{Model Hyperparameter} We perform heuristic-guided searches for tuning hyperparameters that maximize metrics on the validation set. The best hyperparameters picked for each model are listed in Table \ref{tab:hyper}. We set the ratio between the output of MPNN and the structural feature encoder as a learnable parameter.

\paragraph{Training Process} We use Adam \cite{kingma2014adam} as the default optimizer for model parameters with a learning rate $0.0005-0.01$ and set the maximum epoch of 500. Each positive edge is paired with a randomly picked negative edge, except for \texttt{vessel}. Pairwise structural features are computed on the fly from a pair of Bloom signatures.

\paragraph{Selection of Hashing Functions}
The hash function used in Eq. (\ref{eq:bs}) only requires the property of pairwise independence, which has been shown to be easy to implement and efficient in practice \cite{network_bloom}. We use the Murmur3 hash function\footnote{\url{https://github.com/Baqend/Orestes-Bloomfilter\#hash-functions}} in the experiments, as it offers a uniform distribution while being fast. Cryptographic hash functions such as SHA-256 offer even better uniformity albeit slightly slower, so they may potentially provide better quality with \proj. Generally, any good quality hash function while relatively fast can be adopted, and a list of good candidates can be found at \href{https://github.com/rurban/smhasher}{SMhasher}.

\begin{table*}[htp]
\centering
\caption{Hyperparameters used for MPNN with \proj.}
\label{tab:hyper}
    \begin{tabular}{lcccccc}
        \toprule
        \textbf{Dataset} & backbone & \#layers &\#hops $k$ & hashing dim $n$ & learning rate & dropout\\
        \midrule
        \texttt{\texttt{collab}} & GCN & 3 & 2 & $[2048,8192]$ & 0.001 & 0 \\
        \texttt{\texttt{ddi}} & SAGE & 2 & 2 & $[4096,8192]$ & 0.005 & 0.5\\
        \texttt{\texttt{ppa}}  & GCN & 3 & 2 & $[2048,8192]$ & 0.01 & 0\\
        \texttt{\texttt{citation2}} & GCN & 3 & 2 & $[2048,8192]$ & 0.0005 & 0\\
        \texttt{\texttt{vessel}} & SAGE & 3 & 2 & $[2048,4096]$ & 0.01 & 0\\
      \bottomrule
    \end{tabular}
\end{table*}

\section{Deferred Proofs}
\subsection{Error Bounds for Pairwise Structural Feature Estimation from \proj{s} \label{appx:eb}}
\paragraph{Problem Setting} Given a $N$-dimensional binary vector $\vu \in \{0,1\}^N$, \proj reduces it to a $n$-dimensional binary vector $\vs_u \in \{0,1\}^n$. Let $n = \tau m \sqrt{m\ln\left(\frac{2}{\delta}\right)}$, and $m$ is the number of non-zero values (sparsity) in $\vu$ with probability at least $1-\delta'$. It randomly maps each bit position $\{i\}_{i=1}^N$ to an integer $\{j\}_{j=1}^n$ as shown in Eq. (\ref{eq:bs}). To compute the $j$-th bit of $\vs_u$, it checks which bit positions have been mapped to $j$, computes the \texttt{bitwise - OR} of the bits located at those positions and assigns it to $\vs_u[j]$.

\begin{lemma}\label{lemma:1}
\begin{align}
    \mathbb{E}\left[\frac{|\vs_u|}{n}\right]&= 1-\rho^{|\vu|}, \text{~where~} \rho=1-\frac{1}{n} \in (0,1),\\
        \nonumber
    \mathbb{E}\left[\frac{\langle \vs_u, \vx_v\rangle}{n}\right]&=(1-\rho^{|\vu|})(1-\rho^{|\vv|})+\rho^{|\vu|+|\vv|}\left[\left(\frac{1}{\rho}\right)^{\langle \vu,\vv \rangle}-1\right]\\
    &=1-\rho^{|\vu|}-\rho^{|\vv|}+\rho^{|\vu|+|\vv|-\langle \vu,\vv \rangle}.
\end{align}

\end{lemma}
\begin{lemma}\label{lemma:2}
    Given $n \geq cm^{\frac{3}{2}}$, with probability at least $1-\delta$, it holds that 
    \[\big||\vs_u| - \mathbb{E}[|\vs_u|]\big|<\frac{1 + \sqrt{(c^2 + c)\log \frac{2}{\delta}}}{c^2} \sqrt{m}.\]
\end{lemma}

\begin{proof}
Given a measurable space $(\Omega, \mathcal{F})$ with $\Omega = \{0, 1, ..., n\}$ and corresponding $\sigma$-algebra $\mathcal{F}$, consider a sequence of random variables $X_1, X_2, ..., X_m$. Here, $X_k$ represents the count of occupied bins after $k$ ball throws. As the processes are independent, the probability for the $k+1$-th ball to land in an empty bin is $1-\frac{X_k}{n}$. This implies $X_{k+1}=X_k+Z_{k+1}$, where $Z_{k+1}$ follows a Bernoulli distribution with parameter $1-\frac{X_k}{n}$ (noted as $\text{Ber}(1-\frac{X_k}{n})$), and we initialize with $X_0 = 0$. Further, define $Y_k = \frac{X_k - n}{(1-\frac{1}{n})^{k-1}}$ and $Y_1 = 1-n$, we observe that
\begin{equation}
\begin{aligned}
    \mathbb{E}(Y_{k+1}|\sigma(Z_1, ..., Z_k)) &= \frac{X_k - n}{(1 - \frac{1}{n})^k} + \frac{1 - \frac{X_k}{n}}{(1 - \frac{1}{n})^k} + \mathbb{E}\left(\frac{Z_{k+1}}{(1 - \frac{1}{n})^k}\right) \\
    &= \frac{(1-\frac{1}{n})X_k+1-n}{(1-\frac{1}{n})^k} = \frac{X_k - n}{(1-\frac{1}{n})^{k-1}}=Y_k.
\end{aligned}
\end{equation}
Thus, $(Y_{k+1})$ form a martingale adapted to filteration $(\sigma(Z_1, ..., Z_{k}))$. From the Azuma-Hoeffding inequality, we have
\begin{equation}
    \mathbb{P}[|Y_m-Y_1|\geq \epsilon] \leq 2\exp\left(\frac{-\epsilon^2}{2\sum_{k=1}^{m-1}c_k^2}\right),
\end{equation}
where $c_k = \frac{1}{(1-\frac{1}{n})^{k-1}}\geq |Y_{k+1}-Y_k|$.

Consider $\sum_{k = 1}^{m-1}c_k^2$,
\begin{equation}
\begin{aligned}
    \sum_{k = 1}^{m-1}c_k^2 = & \sum_{k = 1}^{m-1} \frac{1}{(1 - \frac{1}{n})^{k-1}} \leq \int_{0}^{m-1} e^{x\log\left(\frac{1}{1 - \frac{1}{n}}\right)} dx \\
     = & \frac{1}{\log(1 + \frac{1}{n-1})} \cdot \left[\left(1 + \frac{1}{n-1}\right)^{m-1} - 1\right] \\ \stackrel{(i)}{\leq} & \left(n - \frac{1}{2}\right) \left[\left(1 + \frac{1}{n-1}\right)^{m-1} - 1\right],
\end{aligned}
\end{equation}
where $(i)$ is from $\frac{2x}{2 + x} \leq \log(1 + x), \forall x \geq 0$.

Let $m = \mathcal{O}_n(n^{\frac{2}{3}})$, specifically, $n \geq cm^{\frac{3}{2}}$, to have the concentration hold for $Y_k$, we need to make sure that
\begin{equation}
\begin{aligned}
    &2\exp\left(\frac{-\epsilon^2}{2\sum_{k=1}^{m-1}c_k^2}\right) \leq \delta \\
    \Leftrightarrow & \frac{\epsilon^2}{2\sum_{k=0}^{m-1}c_k^2} \geq \log\frac{2}{\delta}\\
    \Leftarrow & \epsilon^2 \geq \left(\left(1+\frac{1}{n-1}\right)^{(m-1)}-1\right)\left(n-\frac{1}{2}\right)\log\frac{2}{\delta}\\
    \Leftarrow & \epsilon^2 \geq \left(m + \frac{1}{c}\sqrt{m} + o(\sqrt{m})\right)\log\frac{2}{\delta} \Leftarrow \epsilon \geq \sqrt{\left(1 + \frac{1}{c}\right)m\log\frac{2}{\delta}}.
\end{aligned}    
\end{equation}

From the concentration of $Y_k$, we further derive the concentration property of the number of non-empty bins, i.e. $X_m - X_0$. Since $|Y_m - (1-n)| \leq \epsilon~w.h.p.~Y_m \sim (1-n)\pm \epsilon$, we have
\begin{equation}
\begin{aligned}
    &|X_m - X_0 - m|\\
    = & \left|\left(1-\frac{1}{n}\right)^{m-1}Y_m + n -m \right| \\
    = & \left|\left(1-\frac{1}{n}\right)^{m-1}[(1-n)\pm \epsilon] + n - m \right| \\
    = & \left|\left(1-\frac{m}{n}\right) - \left(1-\frac{1}{n}\right)^{m}\left(1 \pm \frac{\epsilon}{n-1})\right)\right| \cdot n \\
    = & \resizebox{.95\hsize}{!}{$\left|1 - \frac{m}{n} - \left(1 - \frac{m}{n} + \frac{m^2}{2n^2}\right)\left(1 \pm \frac{\sqrt{\left(1 + \frac{1}{c}\right)\log \frac{2}{\delta}}\cdot\sqrt{m}}{n-1}\right) + o\left(\frac{\sqrt{m}}{n}\right)\right| \cdot n$}\\
    \leq & \left(\frac{m^2}{2n^2} + \sqrt{\left(1 + \frac{1}{c}\right)\log \frac{2}{\delta}}\cdot\frac{\sqrt{m}}{n}\right) \cdot n \leq \frac{1 + \sqrt{(4c^2 + 4c)\log \frac{2}{\delta}}}{2c^2} \sqrt{m}.
\end{aligned}    
\end{equation}
\end{proof}

\begin{lemma}\label{lemma:3}
With probability at least $1-\delta$, it holds that
\[
\big||\mathcal{N}(u)| - \hat{n}_u\big| < \left(\frac{1}{2c^2\sqrt{\log \frac{2}{\delta}}} + \frac{\sqrt{c^2 + c}}{c^2}\right)\cdot \sqrt{m\log\frac{2}{\delta}}.
\]
\end{lemma}
\begin{proof}
From Lemma \ref{lemma:1} and Eq. (\ref{eq:card}), we have $\rho^{|\mathcal{N}(u)|} - \rho^{\hat{n}_u}= \frac{|\vs_u| - \mathbb{E}[|\vs_u|]}{n}$. Let $\gamma = \frac{1 + \sqrt{(4c^2 + 4c)\log \frac{2}{\delta}}}{2c^2}$. From Lemma~\ref{lemma:2}, with probability at least $1 - \delta$, we have
\begin{equation}
\begin{aligned}
    &\left|\rho^{|\mathcal{N}(u)|} - \rho^{\hat{n}_u}\right|= \left|\frac{|\vs_u| - \mathbb{E}[|\vs_u|]}{n}\right|\\
    \leq & \frac{\gamma \cdot \sqrt{m}}{n} = \frac{1}{m} \left(\frac{1}{2c^2\tau\sqrt{\log \frac{2}{\delta}}} + \frac{\sqrt{4c^2 + 4c}}{2c^2\tau}\right).
\end{aligned}    
\end{equation}

From~\citet{pratap2019efficient}, we have
\begin{equation}
\begin{aligned}
    &\frac{1}{2}\left(1 - \rho^{\big||\mathcal{N}(u)| - \hat{n}_u\big|}\right) \leq \frac{1}{m} \left(\frac{1}{2c^2\tau\sqrt{\log \frac{2}{\delta}}} + \frac{\sqrt{4c^2 + 4c}}{2c^2\tau}\right) \\
    \Leftrightarrow & \big||\mathcal{N}(u)| - \hat{n}_u\big| \log \rho \geq \log\left(1 - \frac{2}{m}\left(\frac{1}{2c^2\tau\sqrt{\log \frac{2}{\delta}}} + \frac{\sqrt{4c^2 + 4c}}{2c^2\tau}\right)\right) \\
    \Leftarrow & \big||\mathcal{N}(u)| - \hat{n}_u\big| \leq \frac{2\left(\frac{1}{2c^2\tau\sqrt{\log \frac{2}{\delta}}} + \frac{\sqrt{4c^2 + 4c}}{2c^2\tau}\right)}{m - 2\left(\frac{1}{2c^2\tau\sqrt{\log \frac{2}{\delta}}} + \frac{\sqrt{4c^2 + 4c}}{2c^2\tau}\right)} \cdot \frac{1}{\log \frac{1}{\rho}} \\
    \Leftarrow & \big||\mathcal{N}(u)| - \hat{n}_u\big| < \frac{2\left(\frac{1}{2c^2\tau\sqrt{\log \frac{2}{\delta}}} + \frac{\sqrt{4c^2 + 4c}}{2c^2\tau}\right)}{m - 2\left(\frac{1}{2c^2\tau\sqrt{\log \frac{2}{\delta}}} + \frac{\sqrt{4c^2 + 4c}}{2c^2\tau}\right)} \cdot \tau m \sqrt{m\log\frac{2}{\delta}} \\
    \Leftarrow & \big||\mathcal{N}(u)| - \hat{n}_u\big| < \left(\frac{1}{2c^2\tau\sqrt{\log \frac{2}{\delta}}} + \frac{\sqrt{4c^2 + 4c}}{2c^2\tau}\right)\cdot \tau \sqrt{m\log\frac{2}{\delta}}.
\end{aligned}    
\end{equation}
\end{proof}

\begin{lemma}\label{lemma:4}
With probability at least $1-\delta$, it holds that
\[\left|\langle \vs_u,\vs_v \rangle - \mathbb{E}[\langle \vs_u,\vs_v \rangle]\right|< \sqrt{2\min\{|\mathcal{N}(u)|, |\mathcal{N}(v)|\}\log \frac{2}{\delta}}.\]
\end{lemma}
\begin{proof}
    For any given node neighborhoods $\mathcal{N}(u), \mathcal{N}(v) \in \{0,1\}^N$, we partition $\{1, 2, ..., N\}$ in to four sets: (a) $A = \{j | \mathcal{N}(u)[j] = \mathcal{N}(v)[j] = 1\}$; (b) $B = \{j | \mathcal{N}(u)[j] = 1, \mathcal{N}(v)[j] = 0\}$; (c) $C = \{j | \mathcal{N}(u)[j] = 0, \mathcal{N}(v)[j] = 1\}$; (d) $D = \{j | \mathcal{N}(u)[j] = \mathcal{N}(v)[j] = 0\}$. The random mapping $\mathcal H$ can be viewed as throwing different balls into $n$ bins. Let red balls denote indices in $A$, and blue balls and green balls denote indices in $B$ and $C$, respectively. We say a bin is non-empty if at least one red ball or the same amount of blue and green balls exists. WLOG, we can assume $|B| > |C|$. The process can be simplified as we first throw $|B|$ blue balls. After fixing these balls, we then throw $|A| + |C|$ balls. Given measurable space $(\Omega, \mathcal{F})$ with $\Omega = \{0, 1, ..., n\}$ and corresponding $\sigma$-algebra $\mathcal{F}$, define a sequence of random variables $X_1, X_2, ..., X_{|A| + |C|}$, where $X_k$ represents the count of occupied bins after $k$ ball throws. Since we have Lipschitz condition $|X_{k + 1} - X_k| \leq 1$, and from Azuma-Hoeffding inequality
    \begin{equation}
        \mathbb{P}(\left|\langle \vs_u,\vs_v \rangle - \mathbb{E}[\langle \vs_u,\vs_v \rangle]\right| \geq \epsilon) \leq 2 \exp\left(\frac{-\epsilon^2}{2(|A| + |C|)}\right).
    \end{equation}
    Further let $\delta \geq 2 \exp\left(\frac{-\epsilon^2}{2(|A| + |C|)}\right)$, we have
    \begin{equation}
        \epsilon \geq \sqrt{2(|A| + |C|)\log \frac{2}{\delta}}.
    \end{equation}
    Since $|A| + |C| = \min\{|\mathcal{N}(u)|, |\mathcal{N}(v)|\}$, we have the probability at least $1 - \delta$,
    \begin{equation}
        \left|\langle \vs_u,\vs_v \rangle - \mathbb{E}[\langle \vs_u,\vs_v \rangle]\right| \leq \sqrt{2\min\{|\mathcal{N}(u)|, |\mathcal{N}(v)|\}\log \frac{2}{\delta}}.
    \end{equation}
\end{proof}

To estimate the intersection size $|\mathcal{N}(u)\cap\mathcal{N}(v)|$, we can use their Bloom signatures $\vs_u,\vs_v$ as follows
\[
\hat S(u,v) = \hat n_u + \hat n_v - \frac{\ln \left(\rho^{\hat n_u}+\rho^{\hat n_v}+\frac{\langle \vs_u, \vs_v \rangle}{n}-1\right)}{\ln \rho},
\]
where $\hat n_u = \ln(1-|\vs_u|/n)/\ln\rho$ and $\hat n_v = \ln(1-|\vs_v|/n)/\ln\rho$.

\begin{lemma}
With probability at least $1-3\delta$, it holds that \[\left||\mathcal{N}(u) \cap \mathcal{N}(v)| - \hat S(u,v)\right|<\frac{6\sqrt{m}}{c^2} + \left(5\sqrt{\frac{c+1}{c^3}} + 2\sqrt{2}\right)\sqrt{m\log\frac{2}{\delta}},\]
where $m = \max(|\vu|, |\vv|)$ and $\mathcal{N}(u)$ is represented by the $N$-length binary vector $\vu$.
\end{lemma}
\begin{proof}
From Lemma \ref{lemma:1}, we have 
\[\begin{aligned}
    \langle \vu,\vv \rangle &= |\vu|+|\vv|-\ln\left(\rho^{|\vu|}+\rho^{|\vv|}+\mathbb{E}\left[\langle \vs_u, \vs_v\rangle\right]/n-1\right) / \ln \rho,\\
    \hat S(u,v) &= \hat n_u + \hat n_v - \ln \left(\rho^{\hat n_u}+\rho^{\hat n_v}+\langle \vs_u, \vs_v \rangle/n-1\right) / \ln \rho,
\end{aligned}\]
in which $|\vu|\approx \hat n_u$, $|\vv|\approx \hat n_v$ (Lemma \ref{lemma:3}), and $\mathbb{E}\left[\langle \vs_u, \vs_v\rangle\right]\approx \langle \vs_u, \vs_v \rangle$ (Lemma \ref{lemma:4}) with probability at least $1-\delta$.
In Lemma~\ref{lemma:3}, we have shown that
\begin{equation}
\begin{aligned}
    |\hat n_u - |\vu|| &< \left(\frac{1}{2c^2\sqrt{\log \frac{2}{\delta}}} + \frac{\sqrt{c^2 + c}}{c^2}\right)\cdot \sqrt{m\log\frac{2}{\delta}},\\
    |\hat n_v - |\vv|| &< \left(\frac{1}{2c^2\sqrt{\log \frac{2}{\delta}}} + \frac{\sqrt{c^2 + c}}{c^2}\right)\cdot \sqrt{m\log\frac{2}{\delta}}.
\end{aligned}    
\end{equation}

Then, we have
\begin{equation}
\begin{aligned}
    & |\hat S(u,v) - \langle \vu,\vv \rangle| \\
    \leq & \bigg|\hat n_u + \hat n_v - \ln \left(\rho^{\hat n_u}+\rho^{\hat n_v}+\langle \vs_u, \vs_v \rangle/n-1\right) / \ln \rho \\
    &- \left(|\vu|+|\vv|-\ln\left(\rho^{|\vu|}+\rho^{|\vv|}+\mathbb{E}\left[\langle \vs_u, \vs_v\rangle\right]/n-1\right) / \ln \rho\right)\bigg|\\
    \stackrel{(i)}{\leq} & \resizebox{.95\hsize}{!}{$\frac{2\sqrt{m}}{c^2} + \sqrt{\frac{c+1}{c^3}}\sqrt{m\log\frac{2}{\delta}} + \frac{1}{\ln \rho}\left|\ln\frac{\rho^{|\vu|}+\rho^{|\vv|}+\mathbb{E}\left[\langle \vs_u, \vs_v\rangle\right]/n-1}{\rho^{\hat n_u}+\rho^{\hat n_v}+\langle \vs_u, \vs_v \rangle/n-1}\right|$},
\end{aligned}    
\end{equation}
where $(i)$ is from Lemma~\ref{lemma:3}.

Further,
\begin{equation}
\begin{aligned}
    &\frac{1}{\ln \rho}\left|\ln\frac{\rho^{|\vu|}+\rho^{|\vv|}+\mathbb{E}\left[\langle \vs_u, \vs_v\rangle\right]/n-1}{\rho^{\hat n_u}+\rho^{\hat n_v}\langle \vs_u, \vs_v \rangle/n-1}\right| \\
    \leq & \resizebox{.95\hsize}{!}{$\frac{1}{\ln \rho} \cdot \frac{\left|\rho^{|\vu|}+\rho^{|\vv|}+\mathbb{E}\left[\langle \vs_u, \vs_v\rangle\right]/n - \rho^{\hat n_u}+\rho^{\hat n_v}+\langle \vs_u, \vs_v \rangle/n\right|}{\max\{\rho^{|\vu|}+\rho^{|\vv|}+\mathbb{E}\left[\langle \vs_u, \vs_v\rangle\right]/n-1, \rho^{\hat n_u}+\rho^{\hat n_v}+\langle \vs_u, \vs_v \rangle/n-1\}}$}\\
    \leq & \resizebox{.95\hsize}{!}{$\frac{\frac{2\sqrt{m}}{c^2} + 2\sqrt{\frac{c+1}{c^3}m\log\frac{2}{\delta}} + \sqrt{2m\log\frac{2}{\delta}}}{\max\{\rho^{|\vu|}+\rho^{|\vv|}+\mathbb{E}\left[\langle \vs_u, \vs_v\rangle\right]/n-1, \rho^{\hat n_u}+\rho^{\hat n_v}+\langle \vs_u, \vs_v \rangle/n-1\}}$}\\
    \stackrel{(i)}{\leq} & \frac{4\sqrt{m}}{c^2} + 4\sqrt{\frac{c+1}{c^3}m\log\frac{2}{\delta}} + 2\sqrt{2m\log\frac{2}{\delta}},
\end{aligned}    
\end{equation}
where $(i)$ is from \citet{pratap2019efficient} for reasonable values of $m$ and $\delta$ (Proof of Lemma 12). Gluing the above together, we prove the results, and $c=1$ leads to Theorem \ref{thm:intersec}.
\end{proof}

\subsection{Approximation of Pairwise Heuristics based on Node Neighborhoods\label{appx:approx}}
We prove the existence of the MLP in Theorem \ref{thm:approx} by giving the recipe for constructing such an MLP and showing that it satisfies the specified conditions.

\begin{proof}
The MLP has an input layer of width $2n$, a hidden layer of width $N$, and an output layer of width $1$. It can be formulated as
\begin{align}
    \mathrm{MLP}(\vs_u, \vs_v) = \sigma\left(\begin{bmatrix} \vs_u & \vs_v \end{bmatrix} \mathbf W_1 + \mathbf b_1\right)\mathbf W_2 + \mathbf b_2,
\end{align}
where $\mathbf W_1 \in \mathbb R^{2n \times N}, \mathbf W_2 \in \mathbb R^{N \times 1}, \mathbf b_1 \in \mathbb R^{N}, \mathbf b_2 \in \mathbb R$, and $\sigma$ is the ReLU activation function.

Set the values of weight matrices and bias terms as
\begin{equation}
\begin{aligned}
    &\mathbf W_1[i,j] = \begin{cases}
        1 & \text{if $\mathcal{H}(v_j) = i$ or $\mathcal{H}(v_j) = i+N$} \\
        0 & \text{otherwise}
    \end{cases}, \mathbf b_1 = -\mathbf{1};\\
    &\mathbf W_2[i,0] = d(v_i), \mathbf b_2= \mathbf{0}.
\end{aligned}    
\end{equation}

The intuition is that the hidden layer performs the ``set membership testing'' for all $v \in \mathcal V$. Let $\mathbf a = \sigma\left(\begin{bmatrix} \vs_u & \vs_v \end{bmatrix} \mathbf W_1 + \mathbf b_1\right) \in \mathbb R^{N}$. Then, $\mathbf a_i = 1$ if and only if $\vv_i$ is a positive in both Bloom signatures $\vs_u$ and $\vs_v$, and $\mathbf a_i = 0$ iff $\vv_i$ is a negative in either $\vs_u$ or $\vs_v$.

Consider the expectation of the MLP's output,
\begin{equation}
\begin{aligned}
    &\mathbb E (\mathrm{MLP}(\vs_u, \vs_v)) = \mathbb E \left(\mathbf a \mathbf W_2 + \mathbf b_2\right) \\
    &=\resizebox{.95\hsize}{!}{$\mathbb E\bigg(\sum_{w_i \in \mathcal S_I} d(w_i)\mathbf a_i + \sum_{w_j \in \mathcal S_C} d(w_j)\mathbf a_j +\sum_{w_k \in \mathcal S_{D_v}} d(w_k)\mathbf a_k + \sum_{w_l \in \mathcal S_{D_u}} d(w_l)\mathbf a_l\bigg)$}\\
    &=\resizebox{.95\hsize}{!}{$S(u, v) + \mathbb E\bigg(\sum_{w_j \in \mathcal S_C} d(w_j)\mathbf a_j\bigg)+\mathbb E\bigg(\sum_{w_k \in \mathcal S_{D_v}} d(w_k)\mathbf a_k\bigg) + \mathbb E\bigg(\sum_{w_l \in \mathcal S_{D_u}} d(w_l)\mathbf a_l\bigg)$,}
\end{aligned}    
\end{equation}

where $\mathcal S_I = \mathcal N(u) \cap \mathcal N(v)$, $\mathcal S_C = \mathcal V \backslash (\mathcal N(u) \cup \mathcal N(v))$, $\mathcal S_{D_v}=\mathcal N(u) \backslash \mathcal N(v)$, and $\mathcal S_{D_u}=\mathcal N(v) \backslash \mathcal N(u)$.

Since Bloom filters have no false negatives, $w_i \in \mathcal S_I$ implies $\mathbf a_i = 1$. $\mathbf a_j$ for all $j$ such that $w_j \in \mathcal S_C$ are independent Poisson trials with $\Pr(\mathbf a_j = 1) = p_u p_v$, where $p_u = \big(1 - \frac 1n\big)^{|\mathcal N(u)|}$ is the false positive rate for set membership testing in the Bloom signature $\vs_u$, and similarly for  $w_k \in \mathcal S_{D_v}$ and  $w_l \in \mathcal S_{D_u}$.

By Chernoff bound, we obtain the three following inequalities,
\begin{equation}
\begin{aligned}
    \Pr&\left(\sum_{w_j \in \mathcal S_C} \mathbf a_j \le \bigg(1 + \sqrt{\frac{-3\log \delta}{|\mathcal S_C|p_up_v}}\bigg)|\mathcal S_C|p_up_v\right) \ge 1 - \delta,\\
    \Pr&\left(\sum_{w_k \in \mathcal S_{D_v}} \mathbf a_k \le \left(1 + \sqrt{\frac{-3\log \delta}{|\mathcal S_{D_v}|p_v}}\right)|\mathcal S_{D_v}|p_v\right) \ge 1 - \delta,\\
    \Pr&\left(\sum_{w_l \in \mathcal S_{D_u}} \mathbf a_l \le \left(1 + \sqrt{\frac{-3\log \delta}{|\mathcal S_{D_u}|p_u}}\right)|\mathcal S_{D_u}|p_u\right) \ge 1 - \delta.
\end{aligned}    
\end{equation}

Therefore, by Boole's inequality, we have
\begin{equation}
\begin{aligned}
    \Pr\Bigg(&\mathrm{MLP}(\vs_u, \vs_v) - S(u, v) \le \\
    &\resizebox{0.95\hsize}{!}{$d_\mathrm{max}\Bigg(\left(1 + \sqrt{\frac{-3\log \delta}{|\mathcal S_C|p_up_v}}\right)|\mathcal S_C|p_up_v + \left(1 + \sqrt{\frac{-3\log \delta}{|S_{D_v}|p_v}}\right)|S_{D_v}|p_v + \left(1 + \sqrt{\frac{-3\log \delta}{|S_{D_u}|p_u}}\right)|S_{D_u}|p_u\Bigg)$}\Bigg)\\
    &\ge 1 - 3\delta.
\end{aligned}    
\end{equation}
\end{proof}

Following the same strategy above, we can generalize from neighborhood intersections to their unions and differences. The proof is omitted since it is analogous to the proof of Theorem \ref{thm:approx}.

\begin{theorem}\label{thm:set_union}
Suppose $S(u,v)$ is a neighborhood union-based heuristic with the maximum value of $d(\cdot)$ as $d_\mathrm{max} = \max_{w \in \mathcal V} d(w)$. Let $p_u = (1 - 1/n)^{|\mathcal N(u)|}$ denote the false positive rate of set membership testing in the Bloom signature $\vs_u$ (similarly to $p_v$). Then, there exists an MLP with one hidden layer of width $N$ and activation function $\sigma(x) = \begin{cases}
    0 & \text{if } x < 1 \\
    1 & \text{otherwise}
\end{cases}$ where $x \in \mathbb R$, which takes the Bloom signatures of $u$ and $v$ as input and outputs $\hat{S}(u, v) = \mathrm{MLP}(\vs_u, \vs_v)$, such that with probability 1, $\hat{S}(u, v) - S(u, v) \ge 0$; with probability at least $1 - \delta$, it holds that

\begin{equation}
    \hat{S}(u, v) - S(u, v) \le d_\mathrm{max}\bigg(1 + \sqrt{\frac{-3\log \delta}{|\mathcal S_C|p_up_v}}\bigg)|\mathcal S_C|p_up_v,
\end{equation}
where $\mathcal S_C = \mathcal V \backslash (\mathcal N(u) \cup \mathcal N(v))$.
\end{theorem}

\begin{theorem}
\label{thm:set_diff}
Suppose $S(u,v)$ is a neighborhood difference-based heuristic with the maximum value of $d(\cdot)$ as $d_\mathrm{max} = \max_{w \in \mathcal V} d(w)$. Let $p_u = (1 - 1/n)^{|\mathcal N(u)|}$ denote the false positive rate of set membership testing in the Bloom signature $\vs_u$ (similarly to $p_v$). Then, there exists an MLP with one hidden layer of width $N$ and activation function $\sigma(x) = \begin{cases}
    1 & \text{if } x = 0 \\
    0 & \text{otherwise}
\end{cases}$ where $x \in \mathbb R$, which takes the Bloom signatures of $u$ and $v$ as input and outputs $\hat{S}(u, v) = \mathrm{MLP}(\vs_u, \vs_v)$, such that with probability at least $1 - 2\delta$, it holds that

\begin{equation}
    \resizebox{0.95\hsize}{!}{$\big|\hat{S}(u, v) - S(u, v)\big| \le d_\mathrm{max}\left(\left(1 + \sqrt{\frac{-3\log \delta}{|S_{D_u}|p_v}}\right)|S_{D_u}|p_v + \left(1 + \sqrt{\frac{-3\log \delta}{|S_{D_v}|p_u}}\right)|S_{D_v}|p_u\right),$}
\end{equation}
where $\mathcal S_{D_v}=\mathcal N(u) \backslash \mathcal N(v)$, and $\mathcal S_{D_u}=\mathcal N(v) \backslash \mathcal N(u)$.
\end{theorem}

\begin{figure}
    \centering
    \resizebox{.21\textwidth}{!}{
\begin{tikzpicture}[main_node/.style={circle,draw,minimum size=2em,inner sep=2pt]}]

\node[main_node] (0) at (-0.6571428571428579, 4.785714285714286) {\huge $v_1$};
\node[main_node] (1) at (2.3127056238406416, 3.5555627666977863) {\huge $v_2$};
\node[main_node] (2) at (3.5428571428571427, 0.5857142857142867) {\huge $v_3$};
\node[main_node] (3) at (2.3127056238406443, -2.3841341952692128) {\huge $v_4$};
\node[main_node] (4) at (-0.6571428571428561, -3.614285714285714) {\huge $v_5$};
\node[main_node] (5) at (-3.6269913381263525, -2.3841341952692163) {\huge $v_6$};
\node[main_node] (6) at (-4.857142857142858, 0.5857142857142836) {\huge $v_7$};
\node[main_node] (7) at (-3.6269913381263557, 3.5555627666977876) {\huge $v_8$};

 \path[draw, thick]
(0) edge node {} (1) 
(0) edge node {} (6) 
(0) edge node {} (7) 
(1) edge node {} (2) 
(1) edge node {} (3) 
(2) edge node {} (3) 
(2) edge node {} (5) 
(3) edge node {} (4) 
(4) edge node {} (5) 
(4) edge node {} (7) 
(5) edge node {} (6) 
(6) edge node {} (7) 
;

\end{tikzpicture}}
    \caption{\small 3-regular graph with 8 nodes.} \label{fig:3rg}
\end{figure}
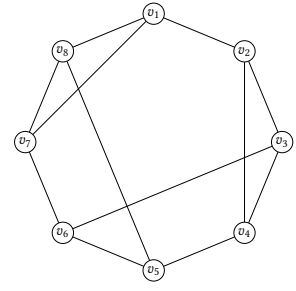

\subsection{Expressiveness Analysis \label{sec:expresiveness}}
The expressive power of link prediction models is measured under Def. \ref{def:mesr} “the most expressive structural representation of link,” which focuses on the model’s distinguishability on automorphic node pairs (rather than graph-level) defined below.

\begin{definition}[Automorphic node pairs]
Let $\mathcal{G}=(\mathcal{V},\mathcal{E})$ be a graph and $e_1=(u_1, v_1), e_2=(u_2, v_2) \in \mathcal{V} \times \mathcal{V}$ be two pairs of nodes. $e_1, e_2$ are automorphic if there exists a mapping $\pi \in \mathbb{A}_G$ (the automorphism group of graph $\mathcal{G}$ defined on its vertex set $\mathcal{V}$ \cite{srinivasan2019equivalence}) such that $\pi \circ e_1 = (\pi(u_1), \pi(v_1)) = (u_2,v_2) = e_2$, denoted as $e_1 \sim_{\mathcal{G}} e_2$. The node automorphism is denoted as $u \sim_{\mathcal{G}} v$ if there exists a mapping $\psi \in \mathbb{A}_G$ such that $u = \psi(v)$. 
\end{definition}

Note that, automorphism between nodes ${u_1 \sim_{\mathcal{G}} u_2, v_1 \sim_{\mathcal{G}} v_2}$ does not imply automorphism between node pairs $(u_1, v_1) \sim_{\mathcal{G}} (u_2, v_2)$, often referred as ``the automorphic node problem''. For example, nodes $w$ and $v$ in Fig. \ref{fig:expressive} are automorphic due to the symmetry, but the pairs $(u,v)$ and  $(u,w)$ are non-automorphic as they have a different number of common neighbors and shortest path distances.

\begin{proposition}
    MPNN with \proj is provably more powerful than the standard MPNN (e.g., GCN/SAGE) and is not less powerful than 3-WL. 
\end{proposition}

\begin{proof}
\textbf{Strictly more expressive than 1-WL.} To show that MPNN with \proj can distinguish between two node pairs whose nodes are automorphic, we consider a 3-regular graph with 8 nodes shown in Fig. \ref{fig:3rg}. Due to the symmetry of this graph, nodes $\{v_1, v_2\}$, $\{v_3,v_4,v_7,v_8\}$, $\{v_5,v_6\}$ are automorphic. For node pairs $(v_1, v_4)$ and $(v_1, v_7)$, they satisfy $v_1 \sim_{\mathcal{G}} v_1$ and $v_4 \sim_{\mathcal{G}} v_7$ but are not automorphic. Assume the exact common neighbors can be recovered from \proj, where $\hat{S}_{\text{CN}}(v_1,v_7)=1$ and $\hat{S}_{\text{CN}}(v_1,v_4)=0$. Then, based on Eq. (\ref{eq:sflp}), our model outputs $\hat{A}_{v_1, v_7} \neq \hat{A}_{v_1, v_4}$ as long as the readout layer MLP ignores the embeddings from MPNN and applies an identity mapping on the pairwise structural features, where such MLP trivially exists. Since our model generalizes from standard MPNN (bounded by 1-WL \cite{xu2019powerful}), it has the same expressiveness of 1-WL without pairwise structural features. This concludes that MPNN with \proj is strictly more powerful than 1-WL.

\begin{figure}[tp]
\captionsetup[subfigure]{font=footnotesize}
\centering
\subcaptionbox{Shrikhande graph}[.235\textwidth]{%
\resizebox{.18\textwidth}{!}{
\begin{tikzpicture}[main_node/.style={circle,draw,minimum size=1em,inner sep=3pt]}]

\node[main_node] (0) at (0.032969597368519565, -1.7463062364848339) {};
\node[main_node] (1) at (1.135400255764967, -0.49252182796840493) {};
\node[main_node] (2) at (1.8095394301652876, 1.4129506271721157) {};
\node[main_node] (3) at (-0.3683929056792228, -0.901183453852294) {};
\node[main_node] (4) at (-2.6781495301940215, 3.9468117329284365) {};
\node[main_node] (5) at (-1.6341798956283073, -1.808739094655081) {};
\node[main_node] (6) at (-3.538803996117302, -1.1354450100159565) {};
\node[main_node] (7) at (1.745692169757823, 3.079540763139289) {};
\node[main_node] (8) at (-4.857142857142858, 1.6337255481075208) {};
\node[main_node] (9) at (-3.9486538957735684, 0.36780738823427406) {};
\node[main_node] (10) at (-4.181487482303318, 3.537804697769319) {};
\node[main_node] (11) at (0.9021178898619002, 2.679006134248212) {};
\node[main_node] (12) at (-3.08062154199353, 4.792892627192561) {};
\node[main_node] (13) at (0.49163089525288495, 4.181353540277399) {};
\node[main_node] (14) at (-1.4140437783547952, 4.857142857142858) {};
\node[main_node] (15) at (-4.792714309465901, -0.03382687359990122) {};

 \path[draw, thick]
(0) edge node {} (1) 
(0) edge node {} (2) 
(0) edge node {} (3) 
(0) edge node {} (4) 
(0) edge node {} (5) 
(0) edge node {} (6) 
(1) edge node {} (2) 
(1) edge node {} (3) 
(1) edge node {} (7) 
(1) edge node {} (8) 
(1) edge node {} (9) 
(2) edge node {} (4) 
(2) edge node {} (7) 
(2) edge node {} (10) 
(2) edge node {} (11) 
(3) edge node {} (5) 
(3) edge node {} (8) 
(3) edge node {} (12) 
(3) edge node {} (13) 
(4) edge node {} (6) 
(4) edge node {} (10) 
(4) edge node {} (12) 
(4) edge node {} (14) 
(5) edge node {} (6) 
(5) edge node {} (11) 
(5) edge node {} (13) 
(5) edge node {} (15) 
(6) edge node {} (9) 
(6) edge node {} (14) 
(6) edge node {} (15) 
(7) edge node {} (9) 
(7) edge node {} (11) 
(7) edge node {} (13) 
(7) edge node {} (14) 
(8) edge node {} (9) 
(8) edge node {} (10) 
(8) edge node {} (12) 
(8) edge node {} (15) 
(9) edge node {} (14) 
(9) edge node {} (15) 
(10) edge node {} (11) 
(10) edge node {} (12) 
(10) edge node {} (15) 
(11) edge node {} (13) 
(11) edge node {} (15) 
(12) edge node {} (13) 
(12) edge node {} (14) 
(13) edge node {} (14) 
;

\end{tikzpicture}}}
\subcaptionbox{4 × 4 Rook’s graph}[.235\textwidth]{%
\resizebox{.18\textwidth}{!}{
\begin{tikzpicture}[main_node/.style={circle,draw,minimum size=1em,inner sep=3pt]}]

\node[main_node] (0) at (-4.857142857142858, 4.857142857142858) {};
\node[main_node] (1) at (-4.802046438410075, -1.0492898913951552) {};
\node[main_node] (2) at (-4.287813196904107, 3.2781954887218046) {};
\node[main_node] (3) at (-4.287813196904107, 0.646616541353382) {};
\node[main_node] (4) at (-0.5779876688967605, 4.389306599832915) {};
\node[main_node] (5) at (1.0749048930867102, 4.857142857142858) {};
\node[main_node] (6) at (-3.0756919847828943, 4.3503202450570875) {};
\node[main_node] (7) at (-0.596353141807688, -0.5034809245335561) {};
\node[main_node] (8) at (1.1300013118194938, -1.06878306878307) {};
\node[main_node] (9) at (-3.0389610389610393, -0.5034809245335561) {};
\node[main_node] (10) at (-1.018759018759019, 2.712893344472292) {};
\node[main_node] (11) at (-1.0003935458480926, 1.094959621275411) {};
\node[main_node] (12) at (0.5606716515807406, 3.2587023113338898) {};
\node[main_node] (13) at (-2.5981896890987812, 2.693400167084379) {};
\node[main_node] (14) at (0.6157680703135258, 0.5686438318017273) {};
\node[main_node] (15) at (-2.5981896890987812, 1.094959621275411) {};

 \path[draw, thick]
(0) edge node {} (1) 
(0) edge node {} (2) 
(0) edge node {} (3) 
(0) edge node {} (4) 
(0) edge node {} (5) 
(0) edge node {} (6) 
(1) edge node {} (2) 
(1) edge node {} (3) 
(1) edge node {} (7) 
(1) edge node {} (8) 
(1) edge node {} (9) 
(2) edge node {} (3) 
(2) edge node {} (10) 
(2) edge node {} (12) 
(2) edge node {} (13) 
(3) edge node {} (11) 
(3) edge node {} (14) 
(3) edge node {} (15) 
(4) edge node {} (5) 
(4) edge node {} (6) 
(4) edge node {} (7) 
(4) edge node {} (10) 
(4) edge node {} (11) 
(5) edge node {} (6) 
(5) edge node {} (8) 
(5) edge node {} (12) 
(5) edge node {} (14) 
(6) edge node {} (9) 
(6) edge node {} (13) 
(6) edge node {} (15) 
(7) edge node {} (8) 
(7) edge node {} (9) 
(7) edge node {} (10) 
(7) edge node {} (11) 
(8) edge node {} (9) 
(8) edge node {} (12) 
(8) edge node {} (14) 
(9) edge node {} (13) 
(9) edge node {} (15) 
(10) edge node {} (11) 
(10) edge node {} (12) 
(10) edge node {} (13) 
(11) edge node {} (14) 
(11) edge node {} (15) 
(12) edge node {} (13) 
(12) edge node {} (14) 
(13) edge node {} (15) 
(14) edge node {} (15) 
;
\end{tikzpicture}}}
\caption{\small The Shrikhande graph and the 4 × 4 Rook’s graph are two non-isomorphic distance regular graphs, where the 3-WL test fails to distinguish them.} \label{fig:srg}
\end{figure}
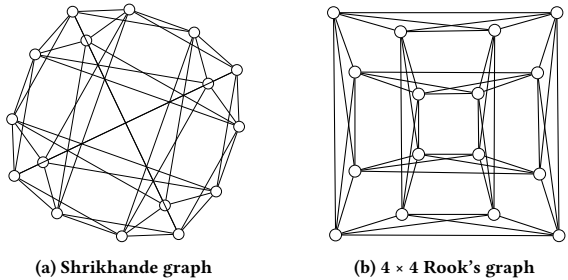

\textbf{Not less powerful than 3-WL.} Consider a pair of strongly regular graphs where the 3-WL test fails \cite{li2020distance}: Shrikhande graph and $4\times4$ Rook’s graph (see Fig. \ref{fig:srg}). Edge representations obtained by MPNN with \proj are conditioned on a pair of nodes, which can achieve the most expressive structural representation of links by Def. \ref{def:mesr}. As Eq. (\ref{eq:ampnn}) shows, each node pair is attached with the joint of pairwise Bloom signatures (equivalently powerful as DE-2 \cite{li2020distance}). As a result, by pooling the obtained representations of all node pairs into a graph-level readout, MPNN with \proj has the capability of distinguishing these two graphs (by different representations for a pair of nodes, where the representations of their common neighbors are different due to the presence of pairwise structural features), which concludes it is not less powerful than 3-WL.
\end{proof}

MPNN with \proj is also more powerful than standard MPNNs in substructure counting. MPNNs (e.g., GCN/SAGE) have been proven to be bounded by 1-WL \cite{xu2019powerful} and unable to count substructures such as triangles \cite{chen2020can}, and thus are consequently incapable of computing common neighbors and other neighborhood overlap-based heuristics such as RA, AA.

\section{Limitations of \proj}
\paragraph{Space and Quality Tradeoff}
\proj is capable of reconstructing any neighborhood intersection/union/difference-based heuristic with guaranteed accuracy (Theorems \ref{thm:approx}, \ref{thm:set_union}, \ref{thm:set_diff}) that covers common link prediction heuristics. Theorem \ref{thm:intersec} indicates the estimation quality depending on the density of node neighborhoods, where it requires more space to maintain high estimation quality for denser graphs. The trade-off between space and estimation error for graphs with different densities is studied and presented in Fig. \ref{fig:random_graph_error}. \proj is superior to the two-step approach of MinHash and HyperLogLog under the same memory budget on both random graphs and real-world networks. To produce the results in Table \ref{table:ogb}, the configuration of hashing and its space overhead of \proj and BUDDY are listed in Table \ref{tab:space}. 

\paragraph{Distance-based Features} \proj is not able to explicitly recover distance features, such as shortest-path distance, but some distance information can be implicitly captured by neighborhood intersection-based features. For example, if the intersection between the 1-hop neighborhoods of two nodes is non-empty, then their shortest path distance is less than or equal to 2.

\paragraph{Update of \proj} When a new node or edge is added, all Bloom Signatures in its neighborhood can be updated without recomputing, since a new neighbor can be added to an existing \proj through a hashing and bit-set operation. When a node or edge is deleted, all Bloom Signatures in its neighborhood need to be recomputed, since the standard \proj does not allow deletions. However, some works, such as the Deletable Bloom filter \cite{rothenberg2010deletable}, allow deletion without recomputing the signature, which is left for future investigation.

\begin{table} [tp]
    \centering
    \caption{The configuration of hashing functions and their space overhead.} \label{tab:space}
    \resizebox{0.48\textwidth}{!}{
    \begin{tabular}{l|c|c|c|c|c}
    \hline
        \multirow{2}{*}{\textbf{Dataset}} & Node Feature & \multicolumn{2}{c|}{Bloom Signature} & \multicolumn{2}{c}{BUDDY}\\ \cmidrule(r{0.5em}){2-6}
        & RAM (GB) & \# of bits $n$ & RAM (GB) & MinHash+HyperLogLog & RAM (GB) \\\hline
        \texttt{collab} & 0.12 & [2048, 8192] & 0.28 & [8192+2048,8192+2048] & 0.56 \\ \hline
        \texttt{ddi} & 0 & [4096, 8192] & 0.0061 & [8192+2048,8192+2048] & 0.01 \\ \hline
        \texttt{ppa} & 0.13 & [2048, 8192] & 0.69 & [8192+2048,8192+2048] & 1.37 \\ \hline
        \texttt{citation2} & 1.50 & [2048, 8192] & 3.49 & [8192+2048,8192+2048] & 6.98 \\ \hline
        \texttt{vessel} & 0.042 & [2048, 4096] & 2.53 & [8192+2048,8192+2048] & 8.43 \\ \hline
    \end{tabular}}
\end{table}